\documentclass[10pt,journal,compsoc]{IEEEtran}

\ifCLASSOPTIONcompsoc
  \usepackage[nocompress]{cite}
\else
  \usepackage{cite}
\fi

\ifCLASSINFOpdf
   \usepackage[pdftex]{graphicx}
\else
   \usepackage[dvips]{graphicx}
\fi
\usepackage{booktabs}
\usepackage{multirow}
\usepackage{multicol}
\usepackage{amsmath}
\usepackage{amssymb}
\usepackage{array}

\ifCLASSOPTIONcompsoc
 \usepackage[caption=false,font=footnotesize,labelfont=sf,textfont=sf]{subfig}
\else
 \usepackage[caption=false,font=footnotesize]{subfig}
\fi

\begin{document}
%
\title{Stable Attribute Group Editing for Reliable Few-shot Image Generation}
%
%
%
%

\author{Guanqi~Ding,
        Xinzhe~Han,
        Shuhui~Wang,~\IEEEmembership{Member,~IEEE,}
        Shuzhe~Wu,
        Xin~Jin,
        Dandan~Tu,
        Qingming~Huang,~\IEEEmembership{Fellow,~IEEE,}

\IEEEcompsocitemizethanks{
\IEEEcompsocthanksitem 
Corresponding author: Shuhui Wang.\\
Guanqi Ding and Xinzhe Han contribute equally to this work.
 \IEEEcompsocthanksitem  G. Ding, X. Han, and Q. Huang are with the School of Computer Science and Technology, University  of  Chinese  Academy  of  Sciences, Beijing 101408, China, and with the Key Laboratory of Intelligent Information Processing, Institute of Computing Technology, Chinese Academy of Sciences, Beijing 100190, China. Q. Huang is also with Peng Cheng Laboratory, Shenzhen 518066, China. 
 E-mail: \{dingguanqi19, hanxinzhe17\}@mails.ucas.ac.cn,
 qmhuang@ucas.ac.cn.
 \IEEEcompsocthanksitem  S. Wang is with the Key Laboratory  of Intelligent Information Processing, Institute of Computing Technology, Chinese Academy of Sciences, Beijing 100190, China, and with Peng Cheng Laboratory, Shenzhen 518066, China.
 E-mail: wangshuhui@ict.ac.cn.
 \IEEEcompsocthanksitem S. Wu, X. Jin, and D. Tu are with Huawei Cloud EI Innovation Lab, China. 
 E-mail: \{wushuzhe2, jinxin11, tudandan\}@huawei.com. \protect}
} 

%
%

\markboth{Journal of \LaTeX\ Class Files,~Vol.~14, No.~8, August~2015}%
{Shell \MakeLowercase{\textit{et al.}}: Bare Demo of IEEEtran.cls for Computer Society Journals}
%



\IEEEtitleabstractindextext{%
\begin{abstract}
Few-shot image generation aims to generate data of an unseen category based on only a few samples. 
Apart from basic content generation, a bunch of downstream applications hopefully benefit from this task, such as low-data detection and few-shot classification. 
To achieve this goal, the generated images should guarantee category retention for classification beyond the visual quality and diversity.
In our preliminary work, we present an ``editing-based'' framework Attribute Group Editing (AGE) for reliable few-shot image generation, which largely improves the performance compared with existing methods that require re-training a GAN with limited data. Nevertheless, AGE's performance on downstream classification is not as satisfactory as expected. This paper investigates the class inconsistency problem and proposes Stable Attribute Group Editing (SAGE) for more stable class-relevant image generation. Different from AGE which directly edits from a one-shot image, SAGE takes use of all given few-shot images and estimates a class center embedding based on the category-relevant attribute dictionary. Meanwhile, according to the projection weights on the category-relevant attribute dictionary, we can select category-irrelevant attributes from the similar seen categories. Consequently, SAGE injects the whole distribution of the novel class into StyleGAN's latent space, thus largely remains the category retention and stability of the generated images. 
Going one step further, we find that class inconsistency is a common problem in GAN-generated images for downstream classification. Even though the generated images look photo-realistic and requires no category-relevant editing, they are usually of limited help for downstream classification. We systematically discuss this issue from both the generative model and classification model perspectives, and propose to boost the downstream classification performance of SAGE by enhancing the pixel and frequency components. Extensive experiments provide valuable insights into extending image generation to wider downstream applications. Code will be available at \url{https://github.com/UniBester/SAGE}.
\end{abstract}



}
\maketitle


\IEEEdisplaynontitleabstractindextext

%
\IEEEpeerreviewmaketitle

\IEEEraisesectionheading{\section{Introduction}\label{sec:introduction}}

\IEEEPARstart{G}{enerative} Adversarial Network (GAN) and its variants~\cite{gan, style, style2} are renowned for the enormous data required to possess satisfying fidelity and variety of generative samples. However, in many real-world vision tasks, such as defect detection~\cite{ren2022state, he2019end} and fine-grained classification~\cite{yu2019hierarchical,lin2017bilinear,van2018inaturalist}, it is difficult to obtain large-scale high-quality training data. Few-shot image generation is to lower the data dependence and generate diverse images from a few samples (say, three to five) of an unknown category. It facilitates more flexible content generation as well as benefits a bunch of downstream applications like low-data detection~\cite{FewShotDetection} and few-shot classification~\cite{FewShotClassification0, FewShotClassification1}. 

Existing few-shot image generation methods can be roughly divided into four types, {\it i.e.}, fusion-based~\cite{gmn, f2gan, matchinggan, lofgan, wavegan}, optimization-based~\cite{figr, dawson}, transformation-based~\cite{dagan, deltagan} and editing-based~\cite{age}. 
Fusion-based methods fuse several input images in a feature space and decode the fused feature back to a realistic image without changing the category. However, given at least two images as inputs, these methods can only generate images similar to the inputs.
Optimization-based methods adopt meta-learning algorithms to generate new images by learning an initialized model and then fine-tuning with each unseen category, but the generated images are blurry and of low quality. 
Transformation-based methods learn intra-category transformations that are applied to one image of the unseen category to generate more images without changing the category. Due to the complexity of intra-category transformations, the end-to-end training is very unstable and leads to crashes in the generated images.
Editing-based methods were first proposed in our previous work AGE~\cite{age}, which models the few-shot image generation as image editing and largely improves the stability needless of re-training a GAN.
AGE divides attributes of the images into category-relevant and category-irrelevant counterparts. The combination of category-relevant attributes, such as the shape of the face and the morphology of the fur, defines the category of the object in an image. Category-irrelevant attributes, such as postures and expressions, reflect differences between images of the same category. Diverse images can be generated by editing the category-irrelevant attributes of the given image without changing its category. 


Compared with previous methods that still struggle with the fidelity and diversity of the generated images, AGE achieves remarkable generation quality, which can largely benefit Artificial Intelligence Generated Content (AIGC) related tasks. 
However, as for the original intention that aims to generate training data for downstream few-shot applications, the improvement is still far from satisfactory. 
The generated images from the GAN models have an unexpected negative effect on some datasets, even if these images are very photo-realistic to humans. Considering that these applications highly rely on category discrimination in the training data, it indicates that the generative model corrupts some subtle but crucial information against real images of a certain category.
We refer to this subtle degradation phenomenon as class inconsistency. 
This paper goes one step further beyond AGE and focuses on generating diverse images of a novel class with category retention. 


\begin{figure}[!t]
\centering
\includegraphics[width=\linewidth]{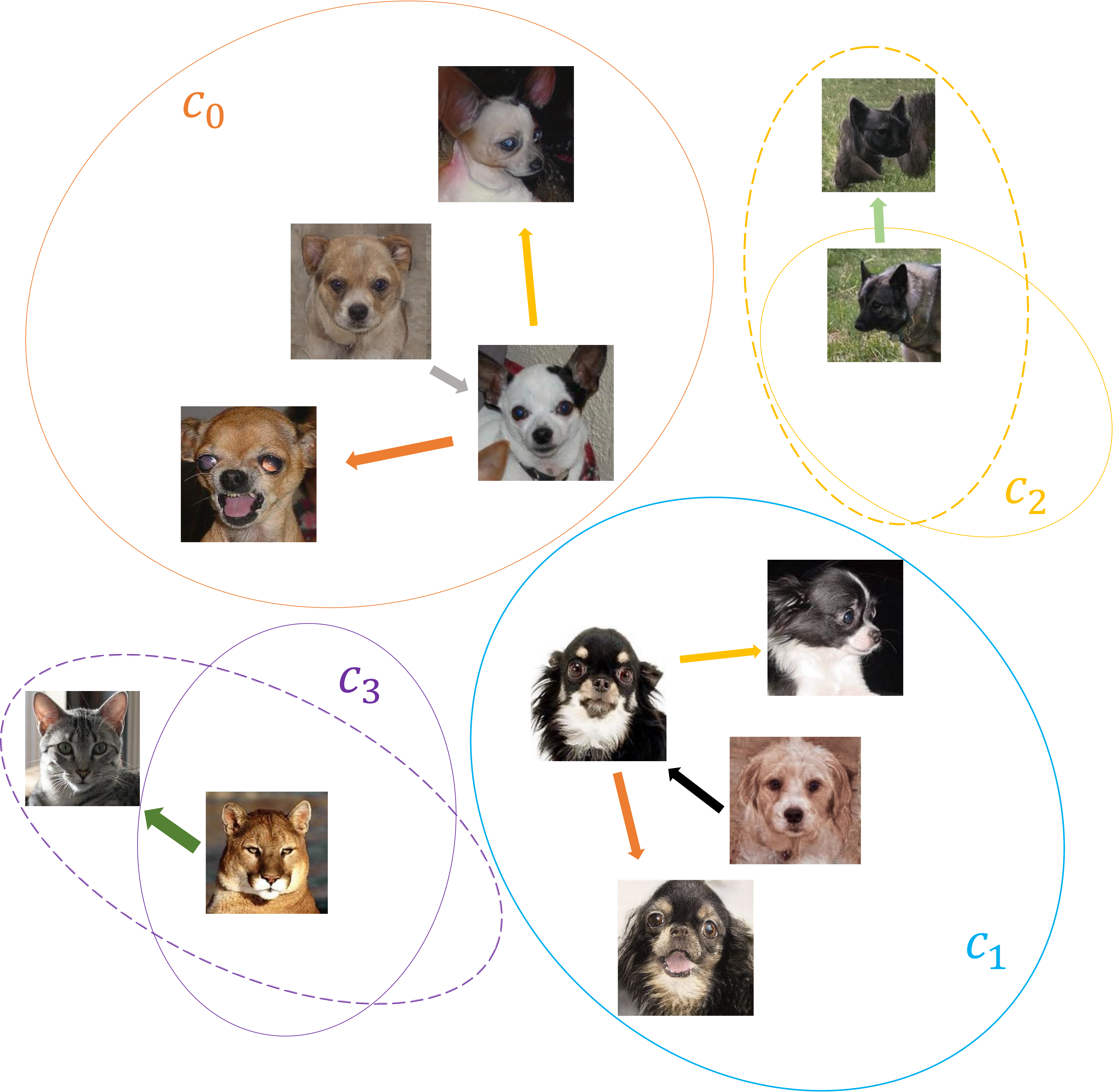}
\caption{Illustration of class distributions in latent space. The true distribution is noted as solid circles while the generative distribution is illustrated as dashed circles. Arrows in different colors denote different specific editing directions. The same attribute editing directions decide the same attributes in all categories. $c_2$ and $c_3$ demonstrate the over-editing and the class inconsistency phenomenons of the editing-based methods such as AGE.} 
\label{principle}
\end{figure}

There are two basic assumptions of the editing-based paradigm AGE. The first is that different samples of one category roughly obey Gaussian distribution in StyleGAN's latent space and the class center, \emph{i.e.}, class embedding, encodes only the category-relevant attributes of a specific category. The second is that the category-irrelevant dictionary is transformable across all categories. Therefore, as demonstrated in $c_0$ and $c_1$ in Fig.~\ref{principle}, any samples can be regarded as editing from the class center with editing directions sampled from the category-irrelevant dictionary. However, some attributes are not category-irrelevant for all categories. For instance, the color of the fur is category-irrelevant for many animals. But for \texttt{puma} ($c_3$ in Fig.~\ref{principle}), it will obviously lead to the change to another species. Besides, AGE directly edits from the given images. If the start point is at an extreme status ($c_2$ in Fig.~\ref{principle}), it tends to show ``over-editing'' as well as unexpected attribute corruptions.
These two shortcomings lead to an offset between the generative distribution and the true class distribution as shown in Fig.~\ref{principle}.

To achieve better category retention on downstream tasks, we propose \textit{Stable Attribute Group Editing (SAGE)}.
It was shown in AGE that the class embedding contains complete category-specific information of one category, which is the best initialization for category-irrelevant editing. 
Injecting the whole novel class distribution into the GAN's latent space will effectively reduce the class inconsistency problem described in Fig.~\ref{principle}.
However, due to the limited number of samples, the class embedding of an unseen category can hardly be obtained by simply calculating the mean latent code as seen categories.
Fortunately, the category-relevant dictionary has encoded almost category-relevant attributes of seen categories, which is complete enough to cover the unique attributes of any related categories. We find that the linear combination of category-relevant attributes can well recover the class-specific features of the unseen classes.
Specifically, SAGE disentangles the category-relevant attributes of a novel class by projecting the available unseen images into the category-relevant dictionary. 
In addition, inspired by the taxonomy in biology~\cite{2010wordnet}, category-irrelevant attributes are more likely to share across the same family or genus, such as different \texttt{orchids} with the same number of petals. 
Species with common category-relevant attributes are more likely to belong to the same family.
Therefore, we further adaptively select category-irrelevant editing directions based on the projection of the category-relevant dictionary when changing the category-relevant features. Accompanied by class embedding, we relocate the whole distribution of the novel class in the latent space, thus fixing the distribution shift in AGE.

To examine SAGE's ability of category retention, we measure the Naive Augmentation Score (NAS)~\cite{cas}, which reflects the generative model's capacity to facilitate downstream classification tasks. Extensive experiments on four representative datasets demonstrate the effectiveness of SAGE on both generative quality and category retention.
Moreover, it shows that taking further use of generated images on downstream tasks not only depend on editing directions but also on the generative model and classifier themselves~\cite{cas}. 
Going one step further on the class inconsistency problem, we explore this issue in-depth and provide an experimental discussion from both generation and classification perspectives. 
In return, we propose a solution to this problem, which enhances the generated content quality by adding discriminatory details from pixel and frequency aspects.
Following our valuable insights, we hope that image generation can be extended to wider downstream applications in the future.

Our contributions can be summarized as follows:

- We propose Stable Attribute Group Editing (SAGE), which highlights category retention in the generation process. SAGE relocates the whole distribution of the novel class in the latent space, thus effectively avoiding category-specific editing and enhancing the stability of the generation.


- We provide extensive experiments on four representative datasets, which suggest that SAGE achieves more stable and reliable few-shot image generation. The facilitation of the downstream classification task is significantly improved compared to the previous methods.

- Apart from the proposed method, we provide an in-depth discussion on the generative class inconsistency problem in downstream classifications.
The experimental analysis demonstrates that making full use of the generative training data requires further investigation of both the generative model itself and the neural classifiers. Based on these findings, we also provide solutions to further boost the downstream classification performance.

This manuscript is an extension of our previous study AGE~\cite{age}. Different from AGE that mainly focuses on generating photo-realistic images, this paper pays attention to the category retention problem, which is of great importance on downstream tasks. 
From methodology, we improve AGE with an adaptive editing strategy and formulate Stable Attribute Group Editing (SAGE). It estimates an adaptive distribution for unseen categories instead of directly editing from given images. Compared with AGE, SAGE largely improves the generative stability and better maintains category discriminatory information for downstream classification.
From experiments, we provide experiments on NABirds~\cite{nabirds} in addition to Animal Faces~\cite{animalfaces}, Flowers~\cite{flowers}, and VGGFaces~\cite{vggfaces}. 
More extensive analysis of SAGE is also provided, including generative distribution comparison, components ablations, and Naive Augmentation Score on downstream classifications.
Apart from the proposed method, we provide a systematic discussion on the class inconsistency problem of generative data augmentations. We experimentally demonstrate that both the generative model and classification model themselves have a great impact on the augmentation in downstream classification besides inappropriate editing. Based on the above analysis, we present solutions to further boost the downstream classification performance by adding information from pixel and frequency aspects. We believe the analysis and solution in this paper may inspire further research.

\section{Related Work}

\noindent\textbf{Few-shot image generation.} Generative models like VAEs~\cite{vae}, GANs~\cite{gan} and Diffusion Models~\cite{diffusion} have achieves remarkable images generation with massive training data. However, diverse image generation with few-shot images is still a challenging task. Existing few-shot image generation methods are mainly based on GANs and can be roughly divided into optimization-based methods, fusion-based methods, and transformation-based methods. Optimization-based methods~\cite{figr, dawson, metafew} combine meta-learning and adversarial learning to generate images of unseen categories by fine-tuning the model. However, the images generated by such methods have poor authenticity. Fusion-based methods fuse the features by matching the random vector with the conditional images~\cite{matchinggan} or interpolate high-level features of conditional images by filling in low-level details~\cite{f2gan,lofgan, wavegan}.
Simple content fusion limits the diversity of generated images. Transformation-based methods \cite{dagan, deltagan} capture the inter-category or intra-category transformations to generate novel data of unseen categories. These works model the transformations from the image differences and may corrupt due to the complex transformations between intra- and inter-category pairs. 
The above existing methods only focus on improving the fidelity and diversity of the generated images.
From our new ``editing-based" perspective~\cite{age}, the intra-category transformation can be alternatively modeled as category-irrelevant image editing based on one sample instead of pairs of samples, and in this paper we pay attention to the category retention problem, which is of great importance in downstream tasks.

\noindent\textbf{Few-shot image-to-image translation.} Few shot image-to-image translation methods map images from one domain to another based on a few images, like category transfer~\cite{animalfaces,munit,snit}, weather transfer~\cite{manifest} and style transfer~\cite{cdimage,ewc,cdone}. 
These methods also focus on the few-shot setting but mainly handle domain transfer rather than object categories.

\noindent\textbf{Image manipulation.} Recent studies have shown that GANs can represent multiple interpretable attributes in the latent space~\cite{sehier}. 
For image editing, some supervised learning methods~\cite{ganalyze, sehier, interfacegan} annotate predefined attributes according to pre-trained classifiers, and then learn the potential directions of the attributes in the latent space. However, they heavily rely on attribute predictors and human annotations.
Some recent work studies unsupervised semantic discovery in GANs. 
The meaningful dimensions can be identified by using segmentation-based networks~\cite{disgan}, linear subspace models~\cite{eigengan}, Principal Components Analysis in the activation space~\cite{RW11-GANSpace}, or carefully designed disentanglement constraints~\cite{sefa,RW11-Hessian,jacobian}.
Different from traditional image editing paradigms, SAGE focuses on attribute factorization for more challenging multi-category image generation.


\noindent\textbf{GAN inversion.}
The task of GAN Inversion~\cite{ganinversion} is projecting real images into their latent representations. 
GAN inversion methods can be divided into optimization-based, encoder-based, and hybrid methods. Optimization-based methods~\cite{lbfgs, i2s, cma} learn the latent code of each sample. They achieve good performance but are slow for inference. Instead of per-image optimization, encoder-based methods~\cite{psp, e4e, restyle, hfgi, hyperstyle} learn an encoder to project images, which are more efficient and flexible but fail to achieve high-fidelity reconstruction. Hybrid methods~\cite{cls, fai} combine the first two kinds of methods and make a compromise between reconstruction quality and editability.
In order to achieve efficient and flexible image editing for few-shot image generation, we choose the encoder-based methods. Specifically, pSp~\cite{psp} is proposed to embed latent codes in a hierarchical manner. e4e~\cite{e4e} analyzes the trade-offs between reconstruction and editing ability. ReStyle~\cite{restyle} finds more accurate latent codes by iterative refinements. Some recent studies~\cite{hfgi, hyperstyle} improve the restoration accuracy by adding bias to the parameters or intermediate features in the generator, which will significantly reduce editability.
\section{Method}

\begin{figure*}[t]
	\centering
	\includegraphics[width=0.95\linewidth]{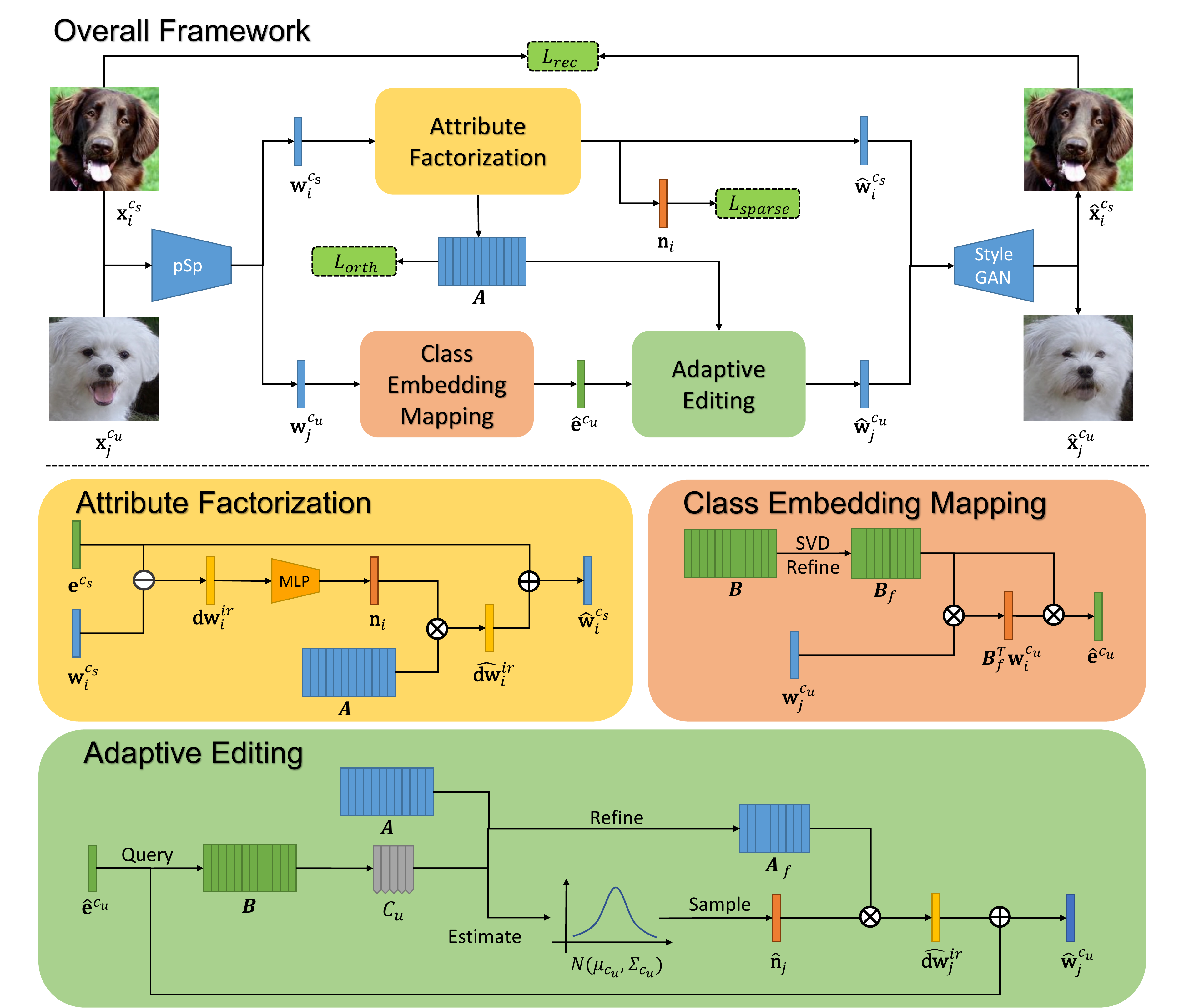}
	
	\caption{The overview of SAGE. SAGE improves class retention by introducing class embedding mapping and adaptive editing.}
	\vspace{-1em}
	\label{structure}
\end{figure*}

\subsection{Preliminaries}
\noindent\textbf{Problem Definition.} First, we clarify the settings of few-shot image generation and introduce the notations we are going to use for the rest of this work.
Few-shot image generation is to train a multi-category image generation model on seen categories and synthesis diverse and plausible images of an unseen category based on $K$ given images. 
$K$ defines a $K$-shot image generation and $K$ is generally small, {\it i.e.}, 10 or 15.
The training set $\mathcal{D}_{train} = \{x_i^{c_s}\}^{S}$ consists of $S$ seen categories and the testing set $\mathcal{D}_{test}= \{x_i^{c_u}\}^{U}$ consists of $U$ unseen categories.

\noindent\textbf{Semantic Manipulation.} Some studies~\cite{ganalyze, sehier, steer} have demonstrated that GANs spontaneously represent multiple interpretable attributes in the latent space, such as color and pose. Different directions in the latent space correspond to different attributes. Given a well-trained GAN, image editing can be achieved by moving the latent code $\mathbf{z} \in \mathcal{Z}$ of an image $x \in \mathcal{X}$ towards a specific direction $\Delta\mathbf{z}$ in the latent space:
\begin{equation}
  \texttt{edit}(G(\mathbf{z}_i)) = G(\mathbf{z'}_i) = G(\mathbf{z}_i + \alpha\Delta\mathbf{z}_i),
  \label{eq:important}
\end{equation}
where $\texttt{edit}(\cdot)$ denotes the editing operation on images, $G(\cdot)$ denotes the generator in GANs, and $\alpha$ stands for the manipulation intensity.

\noindent\textbf{GAN Inversion.} In order to edit a real-world image, its latent code should be obtained at first. GAN inversion is to map images to GANs' latent space: 
\begin{equation}
  \mathbf{z}_i=I(x_i), \ \  \hat{x}_i=G(\mathbf{z}_i).
  \label{eq:gen}
\end{equation}
where $I(\cdot)$ denotes the process of GAN inversion.

\subsection{Stable Attribute Group Editing}
Stable Attribute Group Editing (SAGE) can stably generate images of unseen categories by semantically disentangling attributes and adaptively applying category-irrelevant editing without re-training a GAN model.
The overall framework of SAGE is shown in Figure~\ref{structure}, which consists of two main parts, {\it i.e.}, Attribute Factorization and Stable Generation.

First, to achieve image editing, the latent codes of images should be obtained. We employ pSp~\cite{psp} to encode an image $x_i$ to $\mathcal{W^+}$~\cite{wplus1} space of StyleGAN2~\cite{style, style2}:
 \begin{equation}
  \mathbf{w}_i=\texttt{pSp}(x_i),
  \label{eq:psp}
\end{equation}
where $\mathbf{w}_i\in\mathbb{R}^{18\times512}$ is the corresponding latent vector of $x_i$ in the $\mathcal{W^+}$ space. 

\subsubsection{Attribute Factorization}
In order to achieve category-irrelevant editing, category-relevant attribute directions and category-irrelevant attribute directions in $\mathcal{W}^+$ space should be disentangled.

\noindent\textbf{Category-relevant Attributes}.
Different combinations of category-relevant attributes, which are referred to as class embeddings in $\mathcal{W^+}$, define different categories.  
All samples from the same category can be regarded as moving the same class embedding along different category-irrelevant attribute directions in $\mathcal{W^+}$:
\begin{equation}
  \mathbf{w}_i^{c_s} = \mathbf{e}^{c_s}+\Delta\mathbf{w}_i^{ir},
  \label{eq:important}
\end{equation}    
where $\mathbf{w}_i^{c_s}$ is a sample of $c_s$ in the latent space, $\mathbf{e}^{c_s}$ is the class embedding of $c_s$ and $\Delta\mathbf{w}_i^{ir}$ denotes category-irrelevant editing.

The class embedding should encode the common attributes among all samples from one specific category. We use the mean latent vector $\overline{\mathbf{w}}^{c_s}\in\mathcal{W^+}$ of all samples in a category $c_s$ to represent the class embedding $\mathbf{e}^{c_s}$:
\begin{equation}
    \mathbf{e}^{c_s} = \frac{1}{N_s} \sum_{i=1}^{N_s} \mathbf{w}_i^{c_s},
\end{equation}
where $N_s$ is the number of samples from the category $c_s$.
The dictionary of category-relevant attributes of $S$ seen categories is defined as ${\bf B}=[\mathbf{e}^{c_1},\mathbf{e}^{c_2},...,\mathbf{e}^{c_S}]$.

\noindent\textbf{Category-irrelevant Attributes.} 
Given a latent code $\mathbf{w}_{c_s}$ of any category ${c_s}$, category-irrelevant editing $\texttt{edit}_{ir}(\cdot)$ does not change its category:
\begin{equation}
  \texttt{edit}_{ir}(G(\mathbf{w}_i^{c_s})) = G(\mathbf{w}_i^{c_s} + \alpha\Delta\mathbf{w}^{ir}) = \hat{x}_i^{c_s},
  \label{eq:ir}
\end{equation}
The category-irrelevant directions are common across all known and unknown categories. 
Given a latent code $\mathbf{w}_i^{c_s}$ and its class embedding $e^{c_s}$, category-irrelevant direction sample $\Delta\mathbf{w}_i^{ir}$ can be obtained by:
\begin{equation}
   \Delta\mathbf{w}_i^{ir} = \mathbf{w}_i^{c_s} - \mathbf{e}^{c_s}.
  \label{eq:important}
\end{equation}   

To learn the globally shared category-irrelevant directions, a global dictionary ${\bf A}\in\mathbb{R}^{18\times512\times l}$ and a sparse representation $\mathbf{n}_i \in \mathbb{R}^{18\times l}$ are learned according to $\Delta\mathbf{w}_i^{ir}$ with Sparse Dictionary Learning (SDL)~\cite{sdl, ksvd} as follows:
\begin{equation}
    \min_{\mathbf{n}} \|\mathbf{n}_i\|_0, \ \  \ \ \text{s.t.} \ \  \Delta\mathbf{w}_i^{ir}={\bf A}\mathbf{n}_i, 
  \label{eq:sdl}
\end{equation}
where $\bf{A}$ contains all directions of category-irrelevant attributes, and $\|.\|_0$ is the $L_0$ constraint that encourages each element in $\bf{A}$ to be semantically meaningful. 

In practice, it is optimized via an Encoder-Decoder architecture.
The sparse representation $\mathbf{n}_i$ is obtained from $\Delta\mathbf{w}_i^{ir}$ with a Multi-layer Perceptron (MLP):
\begin{equation}
  \mathbf{n}_i = \texttt{MLP}(\Delta\mathbf{w}_i^{ir}).
  \label{eq:}
\end{equation}

Since the $L_0$ loss is not derivable and it has been proved that $L_1$ is an optimal convex approximation of the $L_0$ norm~\cite{ksvd}, we approximate $L_0$ constraint with $L_1$ using the Sigmoid activation:
\begin{equation}
  L_{\text{sparse}} = \|\sigma(\theta_0\mathbf{n}_i-\theta_1)\|_1,
  \label{eq:reconstruct}
\end{equation}
where $\sigma(\cdot)$ denotes the Sigmoid function. $\theta_0$ and $\theta_1$ are hyper-parameters to control the sparsity. Since the output range from the MLP is very large and unstable, we use Sigmoid activation as a trick to compress the outputs to $[0,1]$. It can better encourage the sparsity of $\mathbf{n}_i$ in our experiment.

The generator is encouraged to generate an image close to the input $x_i$, which is optimized with the $L_2$ reconstruction loss:
\begin{equation}
 L_{\text{rec}} = \|G(\mathbf{e}^{c_s}+{\bf A}\mathbf{n}_i)- x_i^{c_s}\|_2.
  \label{eq:important}
\end{equation}

Moreover, to further guarantee that ${\bf A}\mathbf{n}_i$ only edits category-irrelevant attributes, the embedding of edited images $\mathbf{\hat{w}}^{c_s}$ should have 
\begin{equation}
\begin{aligned}
  {\bf{B}}^T\mathbf{\hat{w}}^{c_s}_i&={\bf{B}}^T\mathbf{e}^{c_s}, \\
  {\bf{B}}^T\mathbf{e}^{c_s}+{\bf{B}}^T{\bf A}\mathbf{n}_i&= {\bf{B}}^T\mathbf{e}^{c_s} ,\\
  {\bf{B}}^T{\bf A}\mathbf{n}_i&=\bf{0}.
  \label{eq:ab}
\end{aligned}
\end{equation}

To ensure the condition of Eq.~\ref{eq:ab}, we formulate an orthogonal constraint between $\bf{A}$ and $\bf{B}$ with:
\begin{equation}
  L_{\text{orth}} = \|{\bf{B}}^T{\bf{A}}\|_F^2,
  \label{eq:oth}
\end{equation} 
where $\|.\|_F^2$ denotes the Frobenius Norm. More intuitively, $ L_{\text{orth}}$ encourages the irrelevance between the category-relevant and the category-irrelevant directions. Regarding $\bf{B}$ as an attribute classifier, we hope the embedding of edited image $\mathbf{w}_i^{c_s}$ has the same responses on category-relevant attributes as the mean vector of its category $\mathbf{e}^{c_s}$.

The overall loss function is
\begin{equation}
  L = L_{\text{rec}} + \lambda_1 L_{\text{orth}} + \lambda_2 L_{\text{sparse}},
  \label{eq:oth}
\end{equation} 
where $\lambda_1$ and $\lambda_2$ are the weights for $L_{\text{orth}}$ and $L_{\text{sparse}}$.

\begin{table*}[t]
\renewcommand\arraystretch{1.5}
\centering
\caption{The quantitative comparison results between SAGE and other methods.}
\begin{tabular}{lccccccccc}
\hline
  \multirow{2}{*}{Method} & \multirow{2}{*}{Settings} & \multicolumn{2}{c}{Flowers}      & \multicolumn{2}{c}{Animal Faces} & \multicolumn{2}{c}{VGGFaces} & \multicolumn{2}{c}{NABirds}                  \\
                          &                           & FID($\downarrow$)        & LPIPS($\uparrow$)        & FID($\downarrow$)         & LPIPS($\uparrow$)        & FID($\downarrow$)       & LPIPS($\uparrow$)      & FID($\downarrow$) & LPIPS($\uparrow$)        \\ \hline
  FIGR~\cite{figr}               & 3-shot                    & 190.12                             & 0.0634                              & 211.54                             & 0.0756                     & 139.83                            & 0.0834                     & 210.75         & 0.0918                       \\
  GMN~\cite{gmn}                 & 3-shot                    & 200.11                             & 0.0743                              & 220.45                             & 0.0868                     & 136.21                            & 0.0902                     & 208.74         & 0.0923                       \\
  DAWSON~\cite{dawson}           & 3-shot                    & 188.96                             & 0.0583                              & 208.68                             & 0.0642                     & 137.82                            & 0.0769                     & 181.97         & 0.1105                       \\
  DAGAN~\cite{dagan}             & 1-shot                    & 179.59                             & 0.0496                              & 185.54                             & 0.0687                     & 134.28                            & 0.0608                     & 183.57         & 0.0967                       \\
  MatchingGAN~\cite{matchinggan} & 3-shot                    & 143.35                             & 0.1627                              & 148.52                             & 0.1514                     & 118.62                            & 0.1695                     & 142.52         & 0.1915                       \\
  F2GAN~\cite{f2gan}             & 3-shot                    & 120.48                             & 0.2172                              & 117.74                             & 0.1831                     & 109.16                            & 0.2125                     & 126.15         & 0.2015                       \\
  LoFGAN~\cite{lofgan}           & 3-shot                    & 79.33                              & 0.3862                              & 112.81                             & 0.4964                     & 20.31                             & 0.2869                     & --             & --                           \\
  WaveGAEN~\cite{wavegan}        & 3-shot                    & 42.17         & 0.3868         & 30.35          & 0.5076 & \textbf{4.96} & 0.3255 & --             & --                           \\
  DeltaGAN~\cite{deltagan}       & 3-shot                    & 78.35        & 0.3487          & 104.62         & 0.4281 & 87.04         & 0.3487 & 95.97          & 0.5136                       \\
  DeltaGAN~\cite{deltagan}       & 1-shot                    & 109.78                             & 0.3912                              & 89.81                              & 0.4418                     & 80.12                             & 0.3146                     & 96.79          & 0.5069                       \\ \hline
  AGE~\cite{age}                                                  & 1-shot                    & 45.96                              & 0.4305                              & 28.04                              & \textbf{0.5575}            & 34.86                             & 0.3294                     & 22.47          & 0.5811                       \\
  SAGE                                                 & 3-shot                    & \textbf{41.35} & 0.4330         & 27.56          & 0.5451 & \textbf{32.89}              &\textbf{0.3314}        & \textbf{19.35} & \textbf{0.5881}              \\
  SAGE                                                 & 1-shot                    & 43.52          & \textbf{0.4392} & \textbf{27.43} & 0.5448 &34.97               &0.3232        & 19.45          & 0.5880                       \\ \hline
\end{tabular}
\label{quantitative_results}
\end{table*}

\subsubsection{Stable Generation}
AGE randomly samples editing directions from $\bf{A}$ and applies it to an unseen category image. The generation process can be expressed as
\begin{equation}
    x_j^{c_u} = G(\mathbf{w}_i^{c_u} + \alpha {\bf A} \tilde{\mathbf{n}}_j),
\label{gaussian}
\end{equation}
where $\tilde{\mathbf{n}}_j$ is the sampled sparse representation and $\mathbf{w}_i^{c_u}$ is the embedding of the given image from unseen category $c_u$.
When $\mathbf{w}_i^{c_u}$ is at the periphery of the distribution of category $c_u$, over-editing can easily occur, \emph{e.g.}, see $c_2$ in Fig.~\ref{principle}. On the other hand, some components in the category-irrelevant attribute dictionary do not adapt to the specific category $c_u$. the sampled sparse representation $\tilde{\mathbf{n}}_j$ may introduce category-relevant editing, \emph{e.g.}, see $c_3$ in Fig.~\ref{principle}.
Therefore, we propose two modules, adaptive editing and class embedding mapping, to achieve stable few-shot generation and better category retention.

\noindent\textbf{Class Embedding Mapping}.
The dictionary $\bf{B}$ is the set of class embeddings of all $S$ seen categories, which contains all category-relevant attribute directions. 
However, class embedding is represented by the mean of the latent codes, and $\bf{B}$ will inevitably contain a small amount of interference from category-irrelevant attributes.
In order to find out the most important directions in $\bf{B}$ and filter out the interference, we conduct singular value decomposition on $\bf{B}$:

\begin{equation}
    {\bf B} = {\bf U}_B {\bf \Sigma} {\bf V}_B^*.
\end{equation}

The top-$t_B$ directions in ${\bf U}_B$ are selected as the final category-relevant attribute dictionary ${\bf B}_f\in\mathbb{R}^{18\times512\times t_{B}}$. 

Given $N_u$ latent codes $\{\mathbf{w}_i^{c_u}\}_{i=1}^{N_u}$ of unseen category $c_u$, the class embedding of $c_u$ can be estimated by the averaged projection from $\mathbf{w}_i^{c_u}$ to ${\bf B}_f$:
\begin{equation}
  \mathbf{\hat{e}}^{c_u}= \frac{1}{N_u} \sum_{i=1}^{N_u} {\bf B}_f{\bf B}_f^{T}\mathbf{w}^{c_u}_i,
  \label{eq:backpro}
\end{equation}
where $\mathbf{\hat{e}}^{c_u}$ encodes the category-relevant attributes and removes most of the category-irrelevant attributes. Editing from $\mathbf{\hat{e}}^{c_u}$ can largely prevent the over-editing phenomenon.



\noindent\textbf{Adaptive Editing}. 
The distribution of category-irrelevant attributes is similar among similar categories, and differs greatly between categories that are quite far away.
Adaptively editing images according to their category allows for better category retention.
Therefore, given a class embedding $\mathbf{\hat{e}}^{c_u}$ of unseen category $c_u$, the top-$t_{C}$ similar seen categories ${\bf C}_u=[c_{s_1}, c_{s_2},...,c_{s_{t_{C}}}]$ are first selected according to the distances between their class embeddings.

In order to find out the category-irrelevant editing directions that best fit the unseen category $c_u$, we first back-project $\Delta\mathbf{w}^{ir}$ into the representation $\mathbf{\hat{n}}$:
\begin{equation}
  \mathbf{\hat{n}}_i={\bf A}^{-1}\Delta\mathbf{w}^{ir}_i,
  \label{eq:backpro}
\end{equation} 
where ${\bf A}^{-1}$ is the pseudo-inverse matrix of ${\bf A}$. Afterward, we count the mean of the absolute value of $|\mathbf{\hat{n}}_i|$ across the similar seen categories ${\bf C}_u$:
\begin{equation}
  \mathbf{\overline{|\hat{n}|}}^{c_u} = \frac{1}{{t_C}} \sum_{t=1}^{t_C} \frac{1}{N_{s_{t}}} \sum_{i=1}^{N_{s_{t}}}  |\mathbf{\hat{n}}_i^{c_{{s_{t}}}}|,
\label{eq:mean_n}
\end{equation} 
where $\mathbf{\overline{|\hat{n}|}}^{c_u}$ can be interpreted as the commonality of the directions across ${\bf C}_u$. For each layer of the $\mathcal{W}^+$ space, we select $t_{A}$ directions from ${\bf A}$ that correspond to the top-$t_{A}$ values in $\mathbf{\overline{|\hat{n}|}}^{c_u}$. The adaptive category-irrelevant dictionary for $c_u$ is ${\bf A}_{c_u}\in\mathbb{R}^{18\times512\times t_{A}}$.


To automatically generate diverse images, we assume that the sparse representation $\mathbf{n}$ of different categories obeys a Gaussian distribution. We estimate the distribution $\mathcal{N}(\mu_{c_u}, \Sigma_{c_u})$ by counting the $\hat{\mathbf{n}}_i$ of all samples in ${\bf C}_u$. We sample an arbitrary $\tilde{\mathbf{n}}_j$ from $\mathcal{N}(\mu_{c_u}, \Sigma_{c_u})$ and apply editing to the estimated class embedding of $c_u$. The manipulation intensity $\alpha$ is introduced to control the diversity of generated images. Given a single image of an unseen category $c_u$, a set of images can be generated by:
\begin{equation}
    x_j^{c_u} = G(\mathbf{\hat{e}}^{c_u} + \alpha {\bf A}_{c_u} \tilde{\mathbf{n}}_j).
    \label{eq:generate}
\end{equation}\
Actually, for one given image, we edit from multiple $\hat{e}^{c_u}$ estimated by different ${\bf B}_f$ with different $t_b$, which can improve the diversity of the generated samples and make the distribution of the generated data closer to the real data.

\begin{figure*}[t]
	\centering
	\includegraphics[width=\linewidth]{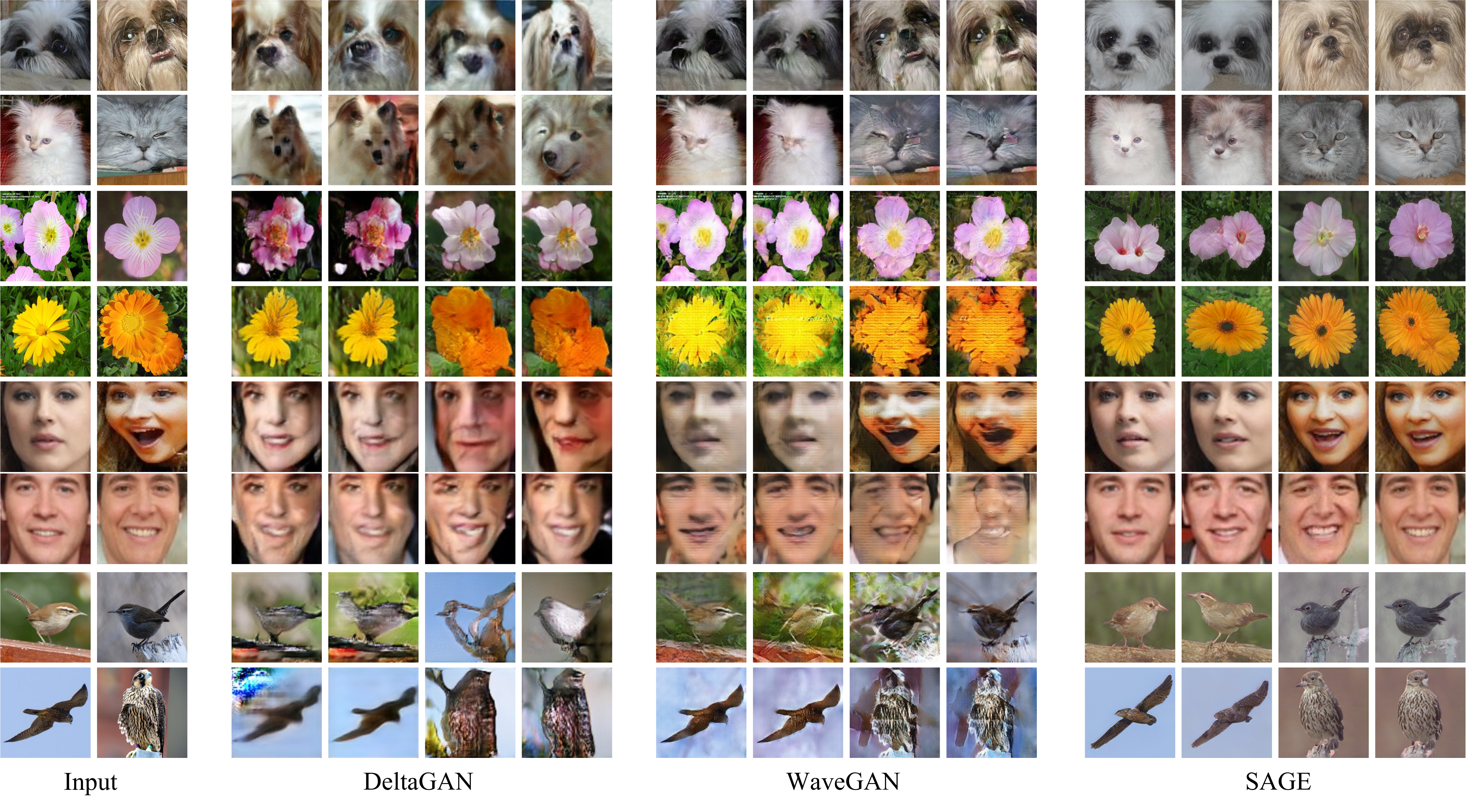}
	\caption{Comparison among images generated by DeltaGAN, WaveGAN, and SAGE on Flowers, Animal Faces, VGGFaces, and NABirds.}
	\label{samples}
\end{figure*}

\section{Experiment}
\label{sec:exp}
\subsection{Implementation Details}
In the training stage, we first train a StyleGAN2~\cite{style2} with images from the seen categories. 
Given a trained GAN, the encoder of sparse representation is a 5-layer multi-layer perceptron with a Leaky-ReLU activation function. The length $l$ of dictionary ${\bf A}$ is set to $100$. 
For more stable and interpretable editing, we divide the 18-layers $\mathcal{W}^+$ space of StyleGAN2 into the bottom layers, middle layers, and top layers, corresponding to 0-2, 3-6, and 7-17 layers respectively. Layers in each group share the same sparse representation $\mathbf{n}$.
We use Adam~\cite{adam} optimizer to train the network for 15000 iterations with a fixed learning rate 0.0001 on four RTX3090 GPUs. 
We set the hyper-parameter $\lambda_1=0.0005$, $\lambda_2=0.005$, $\theta_0=0.5$ and $\theta_1=-1$ to control the influence of loss terms at a similar scale.


\subsection{Datasets}
We evaluate our method on Animal Faces~\cite{animalfaces}, Flowers~\cite{flowers},  VGGFaces~\cite{vggfaces} and NABirds~\cite{nabirds} following the settings in \cite{deltagan}.

\noindent\textbf{Animal Faces}. We select 119 categories as seen categories for training and 30 as unseen categories for testing. 

\noindent\textbf{Flowers}. We split it into 85 seen categories for training and 17 unseen categories for testing. 

\noindent\textbf{VGGFaces}. For VGGFaces~\cite{vggfaces}, we randomly select 1802 categories for training and 572 for evaluation. 

\noindent\textbf{NABirds}. For NABirds~\cite{nabirds}, 444 categories are selected for training and 111 for evaluation.

\subsection{Quantitative Comparison with State-of-the-art}

We evaluate the quality of the generated images based on commonly used FID and LPIPS. FID is used to measure the distance between generated unseen images and real unseen images. We calculate FID between the extracted features of generated unseen images and those of real unseen images. LPIPS is used to measure the diversity of generated unseen images. For each unseen category, the LPIPS score is the average of intra-category pairwise distances among generated images.
Following the former works~\cite{lofgan, matchinggan, deltagan}, the images of each unseen category in $\mathcal{D}_{test}$ are split into $\mathcal{S}_{sample}$ and $\mathcal{S}_{real}$.  We generate 128 images based on $\mathcal{S}_{sample}$ of each unseen category, denoted as $\mathcal{S}_{gen}$. FID and LPIPS are calculated based on the $\mathcal{S}_{real}$ and $\mathcal{S}_{gen}$.

Following one-shot settings in ~\cite{dagan, deltagan}, one real image is used each time to generate adequate number of images for unseen categories. In the three-shot setting, we calculate the mean latent code $\mathbf{\overline{w}}^{c_u}$ of the three unseen samples as the query for similar seen categories. $\mathbf{\overline{w}}^{c_u}$ contains more comprehensive category-relevant attributes of $c_u$ and can achieve more accurate category retrieval on the seen categories. The results of different methods are reported in TABLE~\ref{quantitative_results}, our method achieves significant improvements on both FID and LPIPS in one-shot and three-shot settings. Since there is no need to retrain a GAN, SAGE generates images with higher quality and diversity compared to recently proposed fusion-based methods LoFGAN~\cite{lofgan}, WaveGAN~\cite{wavegan}, and transformation-based methods DeltaGAN~\cite{deltagan}. The employment of class embedding mapping and adaptive editing makes SAGE much more stable than AGE. The estimated distribution prevents over-editing and endows SAGE with more versatile generation ability and remarkable FID and LPIPS gains. 

\subsection{Qualitative Evaluation}
Images generated from SAGE on Animal Faces, Flowers, VGGFaces and NABirds are shown in Fig.~\ref{samples}.
We qualitatively compare our method with the fusion-based methods WaveGAN~\cite{wavegan} and transformation-based methods DeltaGAN~\cite{deltagan}. 

Compared with fusion-based methods that can only interpolate features from the conditioned images, SAGE can produce images with new attributes. Besides, in some cases, WaveGAN does not well disentangle and fuse the features of different images. This can lead to a crash. As shown in Fig.~\ref{samples}, SAGE can generate birds of different orientations, flowers of diverse positions, and humans of different expressions with high fidelity. 

DeltaGAN directly learns the intra-category transformations from different image pairs and  decode these transformations simultaneously. This is very difficult due to the instability of GAN training and the complexity of sample variations. Therefore, the transformations easily lead to category changes and generation crashes. SAGE performs attribute editing in the latent space without GAN re-training. 

\begin{figure}[t]
	\centering
	\includegraphics[width=\linewidth]{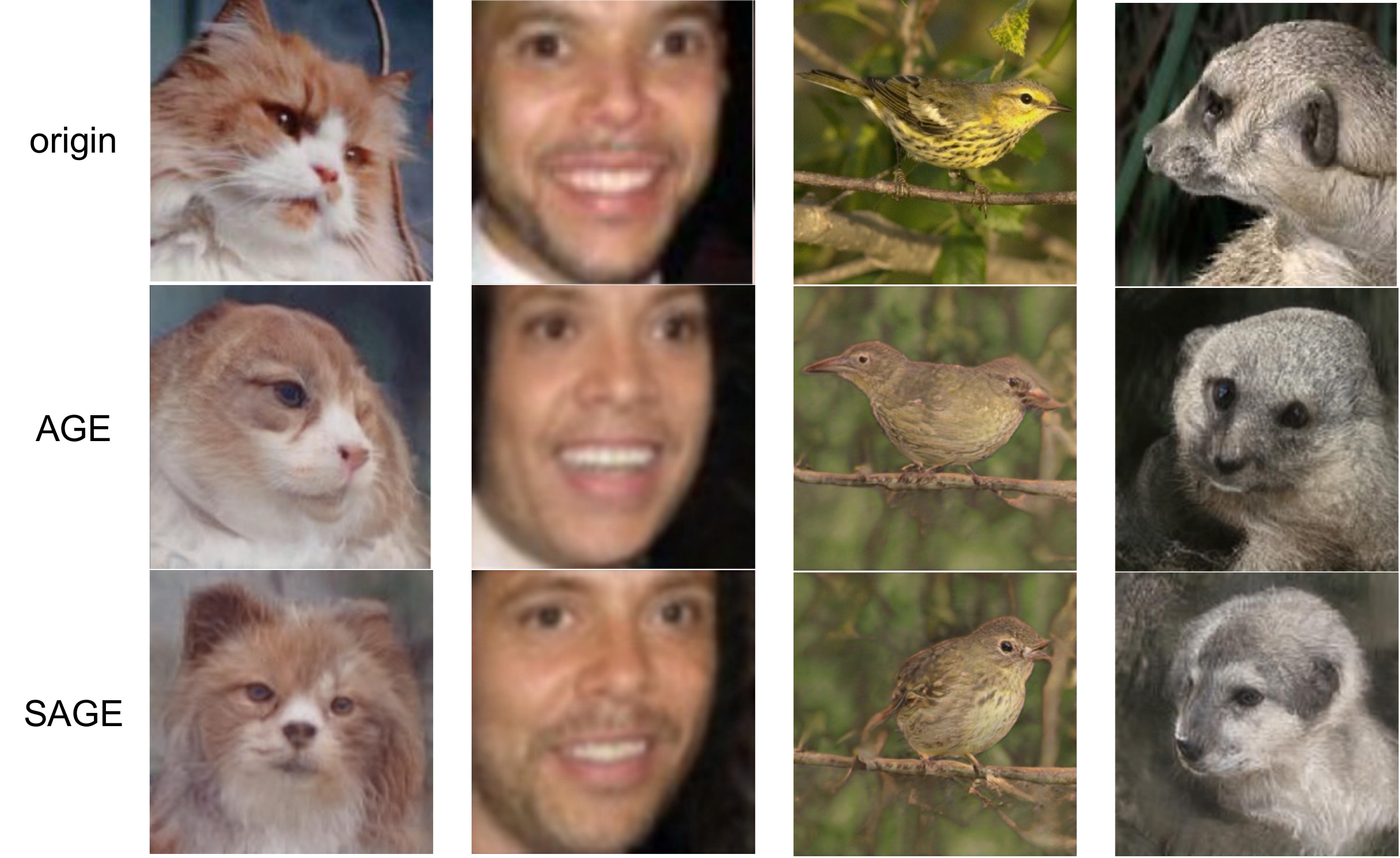}
	\caption{Comparison between images generated by AGE and SAGE. Images in the first two columns that SAGE achieves better category retention. In the last two columns, the editing crash of SAGE is less severe.  }
	\label{age_samples}
\end{figure}

The qualitative comparison between AGE and SAGE is shown in Fig.~\ref{age_samples}. AGE directly samples editing directions from the whole category-irrelevant attribute dictionary, which may lead to category change (\emph{e.g.}, the cat with a dog nose) or over-editing (\emph{e.g.}, the bird with two heads), especially when the starting image is at an extreme posture. SAGE estimates the class embedding and the category-irrelevant attributes for specific unseen categories. 
It achieves few-shot image generation with better diversity, stability and category retention.

\begin{figure}[t]
	\centering
	\includegraphics[width=\linewidth]{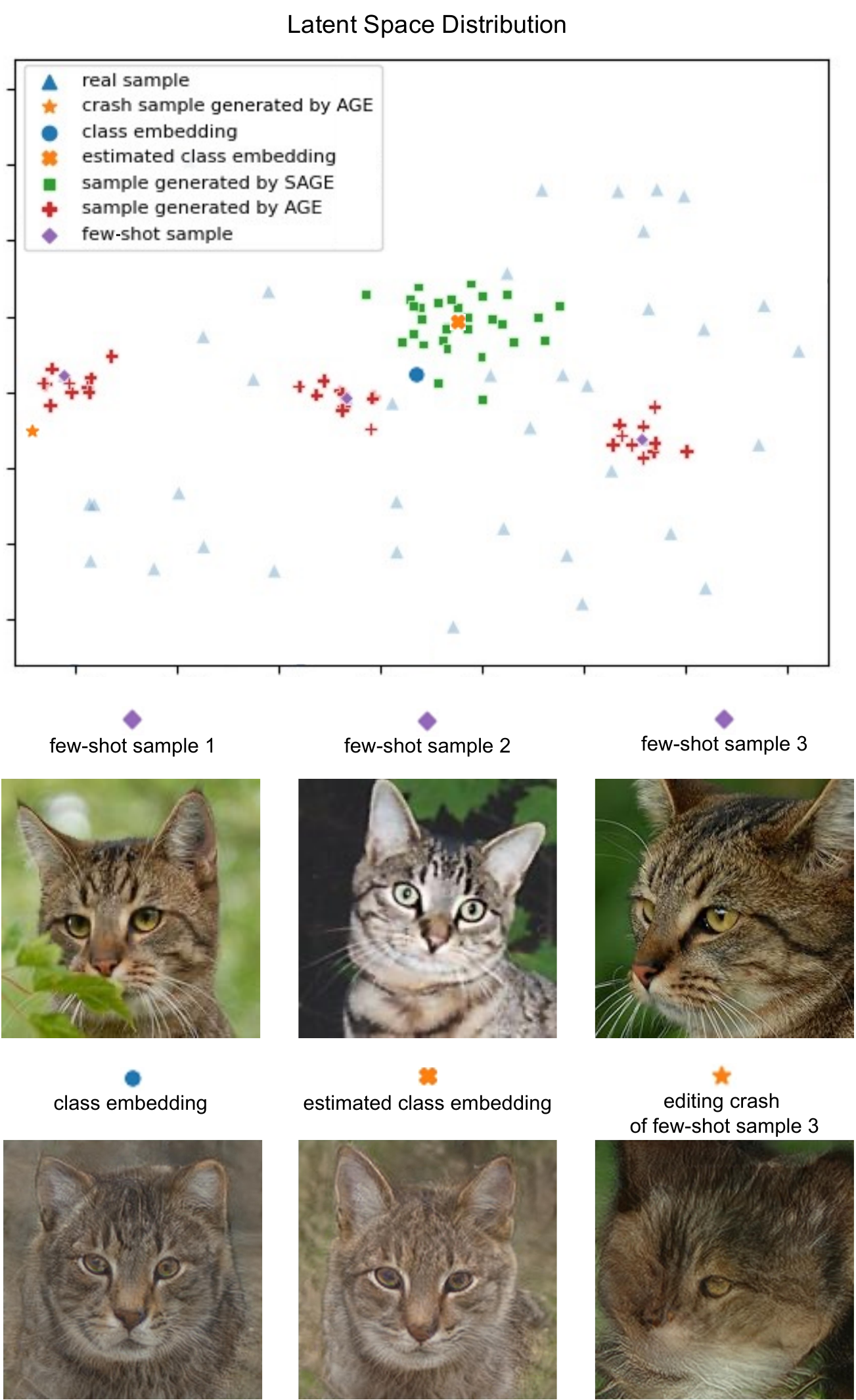}
	\caption{Latent space distribution visualization of real samples and samples generated by AGE and SAGE under the 3-shot setting.}
	\label{distribution_analysis}
\end{figure}

\begin{table}[]
\renewcommand\arraystretch{1.5}
\centering
\caption{Naive Augmentation Score of different kinds of methods on Animal Faces, FLowers, and VGGFaces datasets. VGGFaces and NABirds are marked with * because we randomly select 50 categories from $\mathcal{D}_{test}$ to report NAS.}
\label{nas}
\begin{tabular}{lcccc}
\hline
Method   & Animal Faces     & Flowers          & VGGFaces*     &NABirds*    \\ \hline
DeltaGAN & 47.50\%          & 67.35\%          & 59.27\%     & 39.43\%     \\
WaveGAN  & 51.26\%          & 77.05\%          & 68.94\%      & 46.10\%   \\ \hline
AGE      & 56.46\%          & 75.29\%          & 64.39\%      & 42.81\%    \\
SAGE     & 57.37\% & 77.82\% & 70.53\%  & 45.62\%  \\ \hline
Standard     & 55.38\%          & 79.41\%          & 72.13\%     & 52.60\%     \\ \hline
\end{tabular}
\end{table}

\subsection{Distribution Visualization for Unseen Categories}
In this section, we directly compare the generative distribution of SAGE and AGE in latent space under the 3-shot setting. The real class embedding is the mean vector calculated with all the 700 images of the unseen category in Animal Faces.
The embeddings in $\mathcal{W}+$ space are reduced to 2-D representations with Principle Component Analysis (PCA). The visualizations of real samples, generated samples, class embedding and estimated samples are shown in Fig.~\ref{distribution_analysis}.

As shown in Fig.~\ref{distribution_analysis}, SAGE is not a simple average of the three few-shot samples. Given only three images of diverse postures, the class embedding mapping in SAGE can eliminate the category-irrelevant attributes from the latent code and well retain the category-relevant attributes. Both the estimated class embedding and the corresponding decoded images are very close to the real class embedding calculated with all 700 images.  
In addition, it shows that the distribution of samples generated by SAGE is wider than that of AGE in the latent space. AGE heavily depends on the position of the few-shot samples in the latent space. The generated images are prone to corruption when the reference image is of extreme posture as in sample 3. SAGE relocates the class embedding and adaptively selects the editing directions from the category-irrelevant attribute dictionary, formulating a distribution with a larger editing range. In summary, SAGE can generate images with better stability and diversity.

However, the dynamic range of SAGE generation is still smaller than the real images. According to the scatter plot, the ``editable amplitude'' of different attributes are quite different. SAGE simply assumes a unified intensity $\alpha$ for editing direction sampling since it is almost impossible to make an estimation for each attribute based on only three observations.
Moreover, the category-irrelevant attribute dictionary only encodes the common category-irrelevant attribute shared in all categories. It will be valuable to further capture the class-specific editable attributes or attribute-dependent editing intensity. There is still a large room for improving the editing-based methods on the diversity of the generated images.


\begin{table*}[t]
\renewcommand\arraystretch{1.5}
\centering\caption{The quantitative results of different components of SAGE. AE denotes Adaptive Editing. CME denotes Class Embedding Mapping.}
\begin{tabular}{lllccccccccc}
\hline
Method & \multicolumn{2}{c}{Components}                                                     & Settings & \multicolumn{2}{c}{Flowers}      & \multicolumn{2}{c}{Animal Faces} & \multicolumn{2}{c}{VGGFaces} & \multicolumn{2}{c}{NABirds}  \\
                        & \multicolumn{1}{c}{AE} & \multicolumn{1}{c}{CEM} & 
                                                   & FID($\downarrow$)        & LPIPS($\uparrow$)        & FID($\downarrow$)         & LPIPS($\uparrow$)        & FID($\downarrow$)       & LPIPS($\uparrow$)      & FID($\downarrow$) & LPIPS($\uparrow$)        \\ \hline
AGE                     &                                      &                                             & 1-shot                    & 45.96          & 0.4305          & 28.04          & 0.5575          & 34.86        & 0.3294        & 22.47                      & 0.5811          \\ \hline
SAGE-Q                  &                                      & $\checkmark$                                              & 1-shot                    & 43.52          & 0.4372          & 28.77          & 0.5406          &35.54              &0.3223               & 20.46                      & 0.5790          \\
SAGE-I                  & $\checkmark$                                       &                                             & 1-shot                    & 41.54          & \textbf{0.4479} & 26.05          & 0.5531          &34.77              &\textbf{0.3316}               & 19.68                      & 0.5834          \\
SAGE                    &   $\checkmark$                                     &  $\checkmark$                                             & 1-shot                    & 43.52          & 0.4392          & 27.43          & 0.5448          &34.97              &0.3232               & 19.45                      & 0.5880 \\ \hline
SAGE-Q                  &                                      & $\checkmark$                                            & 3-shot                    & 41.95          & 0.4315          & 2797          & 0.5376          &34.83              &0.3227               & 21.65                      & 0.5707          \\
SAGE-I                  & $\checkmark$                                       &                                             & 3-shot                    & \textbf{40.75} & 0.4410          & \textbf{25.63} & \textbf{0.5593} & \textbf{32.56}              &0.3307               & 22.63                      & 0.5822          \\
SAGE                    &  $\checkmark$                                      &  $\checkmark$                                             & 3-shot                    & 41.35          & 0.4330          & 27.56          & 0.5451          &32.89              &0.3314               & \textbf{19.35}             & \textbf{0.5881}          \\ \hline
\end{tabular}
\label{quantitative_results_ablation}
\end{table*}

\subsection{Data augmentation on Downstream Classification}

To further investigate the improvement in category retention and the contribution of SAGE to downstream tasks, we evaluate and compare the Naive Augmentation Score~\cite{cas} of SAGE against AGE~\cite{age}, fusion-based methods DeltaGAN~\cite{deltagan} and WaveGAN~\cite{wavegan}. Naive Augmentation Score measures the classification accuracy of classifiers trained on augmented data.

Specifically, we first pre-train a Vision Transformer~\cite{vit} network with seen categories. 
Then we split the unseen datasets into $\mathcal{C}_{sample}$, $\mathcal{C}_{val}$ and $\mathcal{C}_{test}$. 
The size of each category in $\mathcal{C}_{sample}$ in different datasets is 10.
The sizes of $\mathcal{C}_{test}$ and $\mathcal{C}_{val}$ vary according to the different amounts of data in different datasets. 
We finetune the pre-trained classifier directly using $\mathcal{C}_{sample}$, which is denoted as "Standard". 
Then we generate 100 images for each category based on $\mathcal{C}_{sample}$ with different methods as augmentation.
We measure NAS on $\mathcal{C}_{test}$, {\it i.e.}, the classification accuracy of the ViT finetuned on data augmented by different methods.

As shown in TABLE~\ref{nas}, SAGE achieves the highest NAS compared to other methods on most of the datasets. 
This indicates that SAGE has better category retention and stronger boost on the downstream classification models.
However, on Flowers, VGGFaces, and NABirds, the NAS of all the few-shot image generation models is not as good as "Standard". 
Compared with Animal Faces, the discrepancy between categories on Flower, VGGFaces and NABirds are much smaller. As mentioned before, such class inconsistency in downstream classifications is very common in the generated image data.
We will further systematically discuss this issue and propose the countermeasure in Section~\ref{exploration}. 

\subsection{Ablation Study}

\begin{table}[t]
\centering
\renewcommand\arraystretch{1.5}
\caption{Ablations of different manipulation intensity $\alpha$.}
\begin{tabular}{@{}cccc@{}}
\hline
$\alpha$ & Accuracy      & FID ($\downarrow$)        & LPIPS ($\uparrow$)                      \\ \hline
Standard   & 55.38\%          & --                                 & --                        \\ 
1.0        & 56.89\% & \textbf{26.23}   & 0.5397     \\
2.0        & \textbf{57.37}\%          & 27.43                              & 0.5448                              \\
3.0        & 56.78\%          & 36.48                              & 0.5474                              \\
4.0        & 56.13\%          & 45.35       & \textbf{0.5531} \\ \hline
\end{tabular}
\label{alpha}
\end{table}

\subsubsection{Manipulation Intensity}
The diversity and quality of generated images are largely controlled by the manipulation intensity $\alpha$.
With the growth of $\alpha$, there is a trade-off between the diversity and quality of the generated images, which are indicated by FID and LPIPS.
Following \cite{age}, the criteria for a good trade-off is to check whether the generated images can improve the unseen categories' few-shot classification performance. 
We test data augmentation for image classification on Animal Faces~\cite{animalfaces}. We randomly select 10, 35 and 100 images for each category as train, validation and test, respectively. A Vision Transformer~\cite{vit} backbone is first initialized using the seen categories, then the model is fine-tuned on the unseen categories. 100 images are generated for each unseen category as data augmentation. The baseline is the ViT model only fine-tuned with 10 real images. The generative process of SAGE follows the 1-shot setting.


According to TABLE~\ref{alpha}, SAGE with $\alpha=2$ achieves the best performance at 57.37\%. 
Compared with AGE in which the best $\alpha = 1.0$, SAGE can sample from the Gaussian distribution with larger covariance, leading to the generation of more diversified images. Since the images are edited from the estimated center of novel class embedding, the probability of over-editing is obviously reduced.

\begin{table}[t]
\renewcommand\arraystretch{1.5}
\centering
\caption{Comparison of Naive Augmentation Score on the impact of different components in SAGE.}
\label{nas_ablation}
\begin{tabular}{lcccc}
\hline
Method & Animal Faces     & Flowers          & VGGFaces*     & NABirds*   \\ \hline
AGE    & 56.46\%          & 75.29\%          & 64.39\%       & 42.81\%   \\
SAGE-Q & 56.75\%          & 75.31\%            & 70.13\%      & 43.66\%    \\
SAGE-I & 56.54\%          & 77.76\%          & 69.85\%     & 45.27\%     \\
SAGE   & \textbf{57.37\%} & \textbf{77.82\%} & \textbf{70.53\%} & \textbf{45.62\%} \\ \hline
\end{tabular}
\end{table}

\subsubsection{Ablations on Different Components of SAGE}
To test the effect of different components in SAGE, we added two settings, {\it i.e.}, SAGE-Q and SAGE-I. In SAGE-Q, we remove the adaptive editing, and select the editing directions from the whole category-irrelevant attribute dictionary. In SAGE-I, we remove the class embedding mapping, and directly edit from the given 1-shot sample. 

As shown in TABLE~\ref{quantitative_results_ablation}, the performance of different settings is basically better than AGE. SAGE-Q provides a good foundation for editing. Especially under the 3-shot setting, the center of class embedding can be better located, resulting in a more stable generation.
Adaptive editing has an even greater impact on FID. Since it selects the feature from relevant seen classes, the overall distributions are closer to the real ones. 

SAGE-Q may sacrifice FID and LPIPS on some datasets compared to SAGE-I, but it guarantees the class information retention for downstream classification tasks.
As the results shown in TABLE~\ref{nas_ablation}, different settings of SAGE achieve consistent improvement on NAS compared to AGE. 
For classification, both SAGE-Q and SAGE-I have improved accuracy compared to AGE.
This indicates that both maintaining the discriminative category information and not introducing category-relevant information from other categories have significant impact on the downstream task.



\begin{figure}[t]
	\centering
	\includegraphics[width=\linewidth]{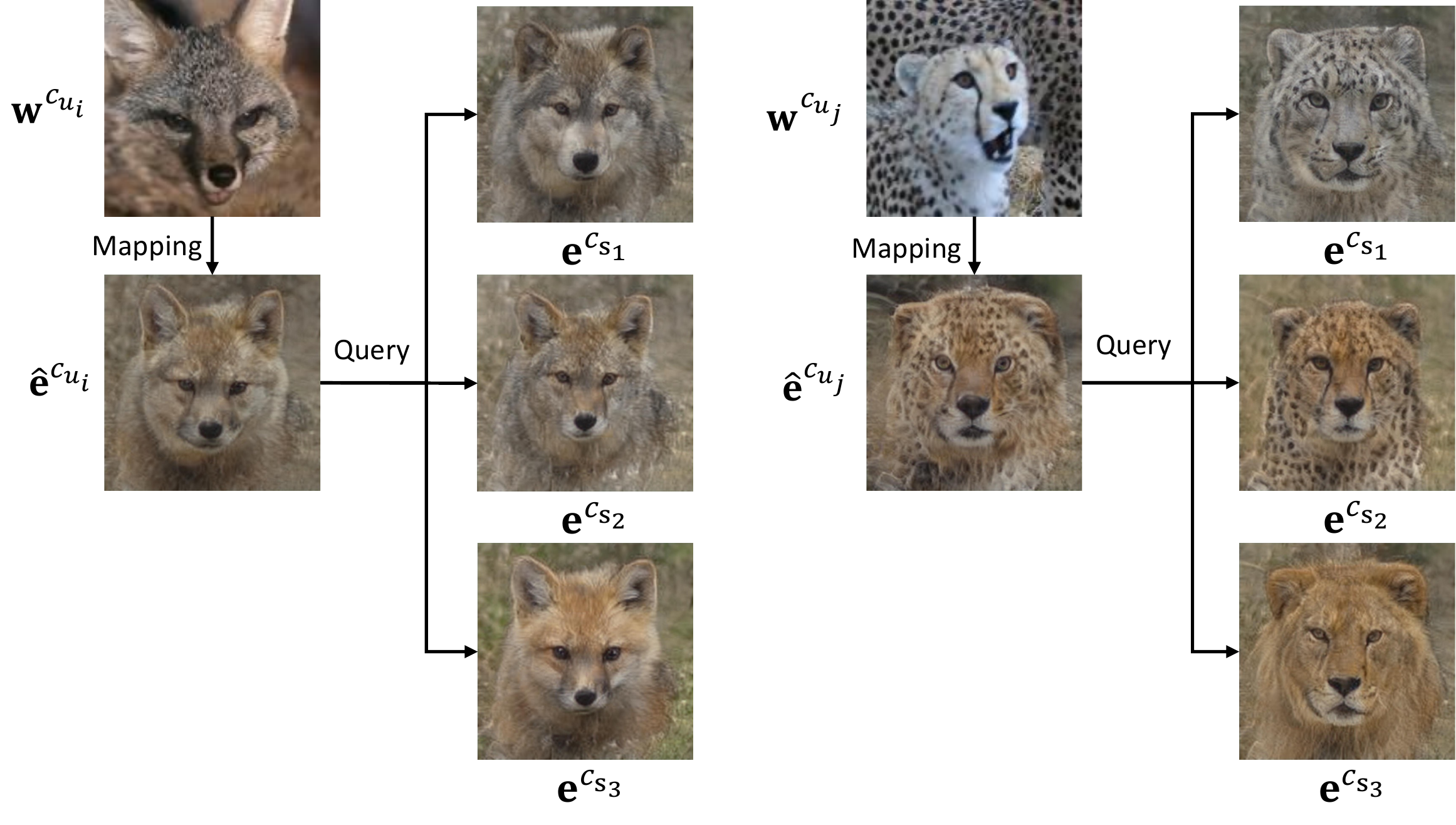}
	\caption{Visualization of the process of adaptive editing. Similar categories can be found by querying the class embeddings.}
	\label{adaptive_editing}
\end{figure}

\begin{table}[t]
\renewcommand\arraystretch{1.5}
\centering
\caption{Ablations of the number of directions in the category-relevant attribute dictionary $t_{B}$ on Animal Faces.}
\label{class_embedding_mapping_ablation}
\begin{tabular}{ccccc}
\hline
$t_{B}$     & 1         & 5          & 10     & 30             \\ \hline
FID($\downarrow$)    & 27.61     & 27.59      & \textbf{27.56}   & 27.63                                         \\
LPIPS($\uparrow$) & 0.5449 & \textbf{0.5462}  &0.5451   & 0.5443    \\ \hline          
\end{tabular}
\end{table}

\subsubsection{Ablations on Category-relevant Attribute Dictionary}
We employ class embedding mapping to overcome the corruption in generated images caused by over-editing. $t_{B}$ controls the number of directions in the category-relevant attribute dictionary ${\bf B}_f$. As shown in Fig.~\ref{class_embedding_mapping_ablation}, smaller $t_{B}$ filters out most interference from the category-irrelevant attributes, making these attributes at a moderate value in the estimated class embeddings. However, it will also inevitably filter out the valuable category-relevant attributes. On the contrary, larger $t_{B}$ will leave a small amount of category-irrelevant attributes in the estimated class embeddings, introducing incorrect category-relevant attributes. We test different $t_{B}$ under the 1-shot setting. As shown in TABLE~\ref{class_embedding_mapping_ablation}, when the value of $t_{B}$ is in a certain interval, it does not have a significant impact on the generated image quality. The best results are obtained between 5 and 10. The class embedding estimated in this interval includes both complete category-relevant information and does not introduce additional category-irrelevant information. In practice, we select $t_{B} = 10$ for similar categories retrieval in adaptive editing.

\begin{figure}[t]
	\centering
	\includegraphics[width=\linewidth]{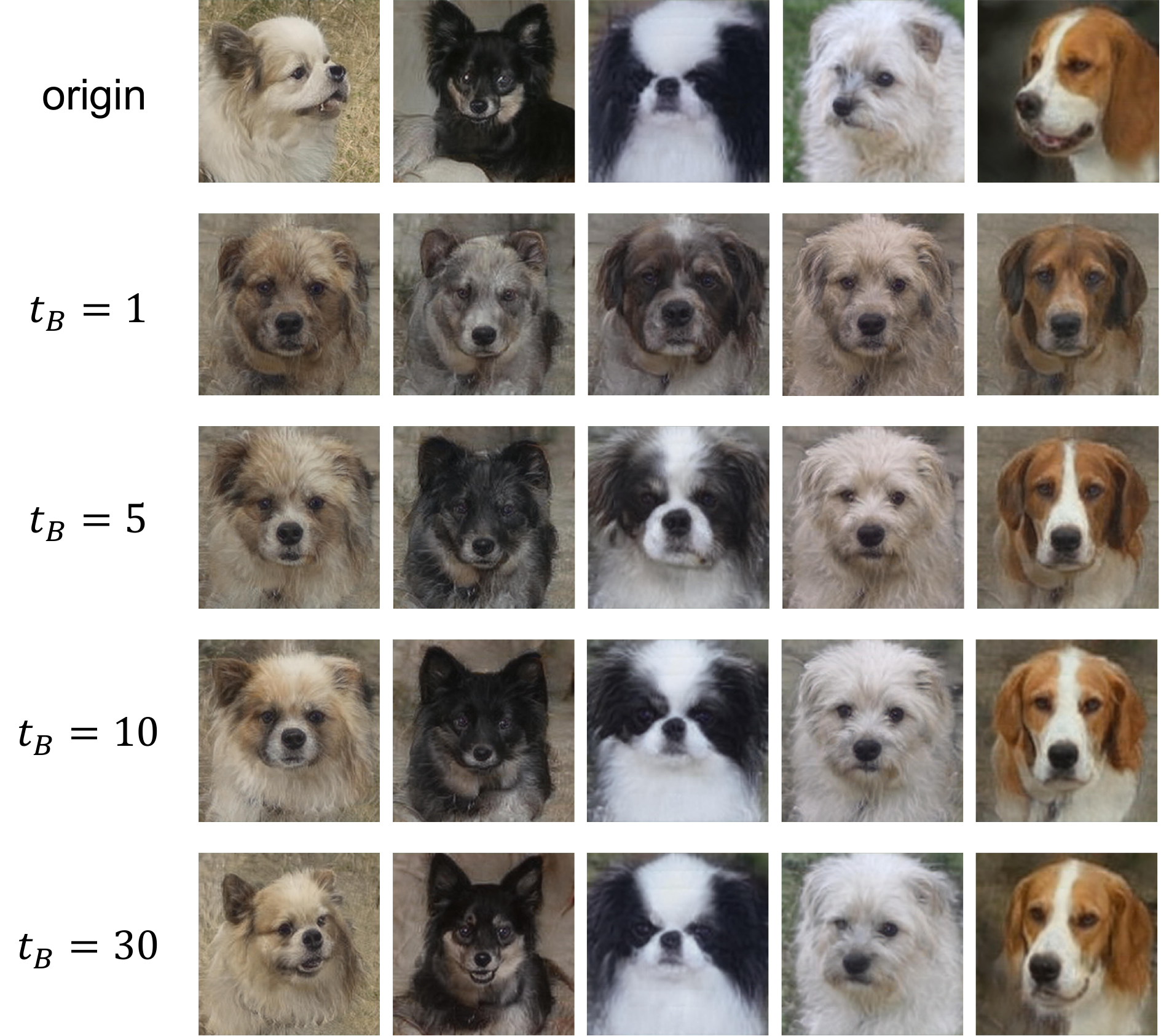}
	\caption{Visualization of the estimated class embeddings with different $t_{B}$.}
	\label{class_embedding_mapping_ablation}
\end{figure}

\subsubsection{Ablations on Category-irrelevant Attribute Dictionary}
Adaptive editing finds $t_{C}$ similar categories of $c_u$ according to its estimated class embedding $\mathbf{\hat{e}}^{c_u}$ and uses the category-irrelevant attribute distribution of these categories to estimate the distribution of $c_u$. 
As shown in Fig.~\ref{adaptive_editing}, similar categories can be found by querying the class embeddings. For example, the similar categories queried with \texttt{cheetahs} were all \texttt{felines}.

The class embedding of $c_u$ inevitably contains biases, since it is estimated with only few-shot images. Retrieving more categories can reduce such biases and comes to a more robust distribution. However, too many retrieved categories will introduce classes that are hardly related to the queried unseen category. 
As shown in TABLE~\ref{adaptive_editing_ablation}, both too small and too large $t_{C}$ will lead to degradation of the generated image quality in both one-shot and three-shot settings. According to this experiment, we set $t_{C} = 30$ for the 1-shot setting and $t_{C} = 20$ for the 3-shot setting.
The 3-shot generation requires fewer similar categories than 1-shot for optimal results. That is because the latent code query in the 3-shot contains more comprehensive category-relevant attributes and retrieves similar categories more accurately.

\begin{table}[t]
\renewcommand\arraystretch{1.5}
\centering
\caption{Ablations of the number of retrieved categories $t_{C}$ on Animal Faces.}
\label{adaptive_editing_ablation}
\begin{tabular}{clccccc}
\hline
\multicolumn{2}{c}{$t_{C}$ }   & 10    & 20      & 30      & 40  & 50  \\ \hline
1-shot  & FID($\downarrow$)    & 26.16    & 26.13   & \textbf{26.05}   & 26.18  & 26.32   \\
        & LPIPS($\uparrow$)    & 0.5522   & 0.5515  & \textbf{0.5531}  & 0.5519                     & 0.5515                     \\ \hline
\multicolumn{1}{l}{3-shot} & FID($\downarrow$)    & \multicolumn{1}{l}{25.72}  & \multicolumn{1}{l}{\textbf{25.63}}  & \multicolumn{1}{l}{25.88}  & \multicolumn{1}{l}{25.66}  & \multicolumn{1}{l}{25.69}  \\
\multicolumn{1}{l}{} & LPIPS($\uparrow$) & \multicolumn{1}{l}{0.5554} & \multicolumn{1}{l}{\textbf{0.5593}} & \multicolumn{1}{l}{0.5543} & \multicolumn{1}{l}{0.5539} & \multicolumn{1}{l}{0.5542} \\ \hline
\end{tabular}
\end{table}

\begin{figure*}[t]
	\centering
	\includegraphics[width=0.9\linewidth]{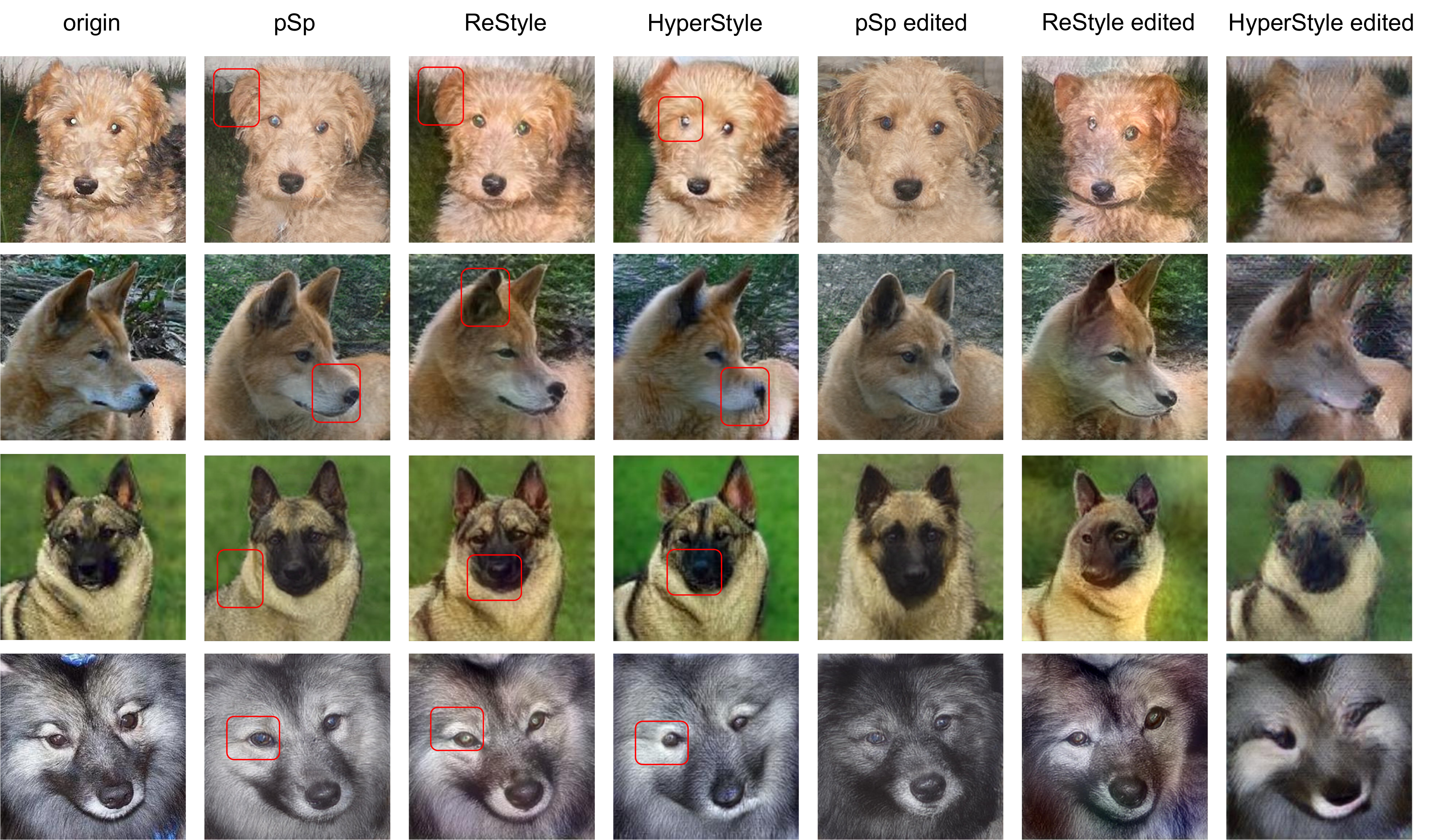}
	\caption{Visualization of the reconstructed images and edited images generated by different inversion methods.}
	\label{inversion}
\end{figure*}

\subsubsection{Ablations on GAN Inversion Methods}
In SAGE, the editing starts from the latent representation of GAN inversion, so the quality of the generated images heavily relies on the quality of GAN inversion. As illustrated in~\cite{e4e}, there is a trade-off between editability and restoration quality in GAN inversion. Some methods~\cite{hyperstyle, hfgi} improve the restoration quality by adding bias to the parameters or intermediate features to the generator, which will significantly reduce editability. We test SAGE upon three different GAN inversion methods pSp~\cite{psp}, HyperStyle~\cite{hyperstyle} and ReStyle~\cite{restyle}. The evaluation of FID, LPIPS and NAS is shown in TABLE~\ref{inversion_comparison}. 
There is a trade-off between restoration quality and editability. As shown in Fig.~\ref{inversion}, GAN inversion methods with better restoration are able to maintain more visual details, but the accompanying degradation of editability will significantly affect the quality of images generated by SAGE. Therefore, we choose pSp as the GAN inversion method for all experiments.

\begin{table}[]
\renewcommand\arraystretch{1.5}
\centering
\caption{Quantitative experiment results of SAGE constructed on different GAN inversion Methods on Animal Faces.}
\label{inversion_comparison}
\begin{tabular}{lccc}
\hline
Method & pSp             & ReStyle & HyperStyle \\ \hline
FID ($\downarrow$)     & \textbf{25.72}  & 32.48   & 39.98      \\
LPIPS ($\uparrow$)  & \textbf{0.5578} & 0.5506  & 0.5521     \\ 
NAS ($\uparrow$)  & \textbf{57.37\%} & 56.91\%  & 47.76\%     \\\hline
\end{tabular}
\end{table}

\section{Discussion on Class Inconsistency}
\label{exploration}
The class inconsistency refers to the corruption of crucial information in the generated images caused by generative models, which has negative effect on the classification model training.
We argue that such degradation in downstream classification tasks not only comes from inappropriate editing.
First, the GAN inversion and the generative process will lead to the lost of some key characteristics for image classification.
Second, existing neural classification methods do not always make predictions with the proper image features, which further enlarges the impact of such generative loss.
In this section, we detail our analysis on the causes of the class inconsistency problem and explore the solution.

\begin{table}[t]
\renewcommand\arraystretch{1.5}
\centering
\caption{Classification accuracy comparison between different inversion methods on Animal Faces.}
\label{inversion_classifcation_comparison}
\begin{tabular}{lcccc}
\hline
Method   & Standard     & pSp          & ReStyle    &HyperStyle        \\ \hline
Accuracy     & 55.38\%          & 49.57\%          & 50.27\%     & 44.26\%         \\ \hline
\end{tabular}
\end{table}

\subsection{Analysis on Generative Models} \label{analyse_GAN}
The generation process of image editing can be divided into two parts, {\it i.e.}, GAN inversion and latent code editing. SAGE mainly focuses on the editing process but the GAN inversion is controlled by the pre-trained inversion method.
We compare the real images and reconstructed images generated by different inversion methods. As shown in Fig.~\ref{inversion}, the overall appearances of the reconstructed images are very similar to the real images. It does not look different by the naked eye in a glimpse. The differences only show in some local details.

\begin{figure}[t]
	\centering
	\includegraphics[width=\linewidth]{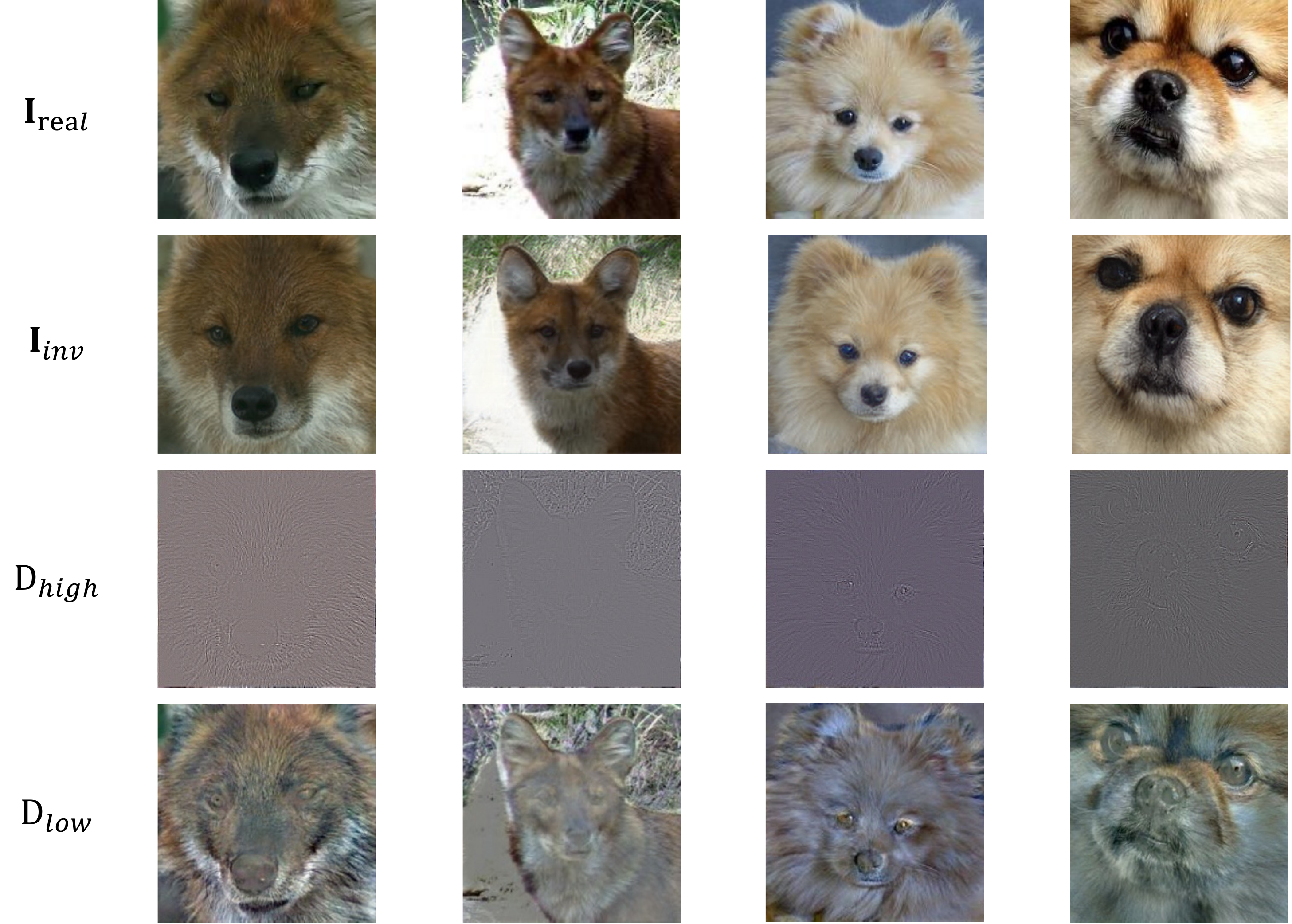}
	\caption{Visualization of the difference between the real images and images inversed by pSp in low frequency and high frequency. As shown in the figure, there is no significant difference between the two at high frequencies, but at low frequencies the difference is quite obvious.}
	\label{frequency_difference}
\end{figure}

To further compare the difference between the real and the generative images, we analyze the generation process from the frequency perspective.
The frequency domain decomposition of an image enables different frequency components in the image to be split and analyzed.
We decompose the high- and low-frequency information with a Gaussian low-pass filter ($\sigma = 5$).
The difference between the reconstructed image and real image is shown in Fig.~\ref{frequency_difference}. 
The major difference lies in low-frequency information, which determines the contour and texture information of the image. 

Although looking similar to humans, the loss of local details greatly influences the accuracy of classification tasks. Using different inversion methods, we inverse the data in $\mathcal{C}_{sample}$ without editing, and train the classifier on these data.
As shown in TABLE~\ref{inversion_classifcation_comparison}, the accuracy with the reconstructed images is lower than the ``Standard" data by over 5\%.
It indicates that the lost local details in GANs contain key discriminative information for the classifier, which comes from both the misalignment between inverse embedding and real embedding and the generative ability of GANs. The information loss in the embeddings will be carried over to the editing stage in both AGE and SAGE. 
Despite of that, SAGE still gains increased diversity of the generated data and improved performance on downstream task compared to directly training on the inversed data. It indicates that SAGE can reduce the gap between the distribution of the generated data and real data.
If we obtain a better inversion method with better reconstruction ability and editability, SAGE will be endowed with a larger potential to further improve downstream classifications with generative augmentation.


\begin{figure}[t]
	\centering
	\includegraphics[width=\linewidth]{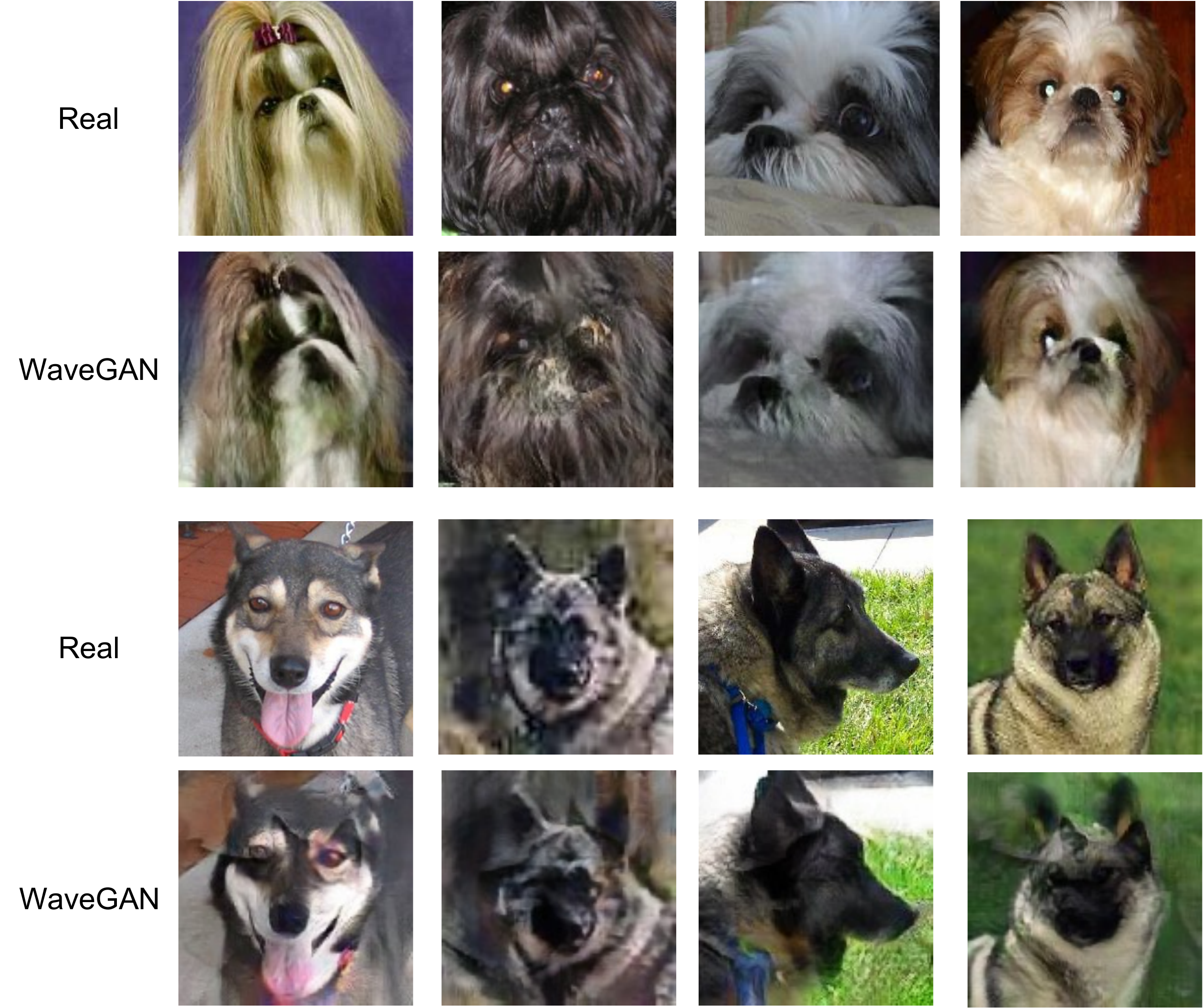}
	\caption{Randomly selected images generated by WaveGAN. A large percentage of WaveGAN-generated images have an overall crash.}
	\label{wavegan}
\end{figure}

\subsection{Analysis on Classifiers} \label{analyse_CLF}
When we re-investigate the data generated by different models, as shown in Fig.~\ref{wavegan}, a large portion of WaveGAN-generated images have an overall crash but the NAS does not show such severe degradation, as shown in TABLE~\ref{nas}. 
We argue that the unsatisfying performance on downstream classification also comes from the classification models. Different from humans who pay attention to the overall information, existing neural classifiers make predictions largely based on the local details, which further protrudes the degradation of local details by the generative augmentation discussed in Section~\ref{analyse_GAN}.

\begin{figure}[t]
	\centering
	\includegraphics[width=\linewidth]{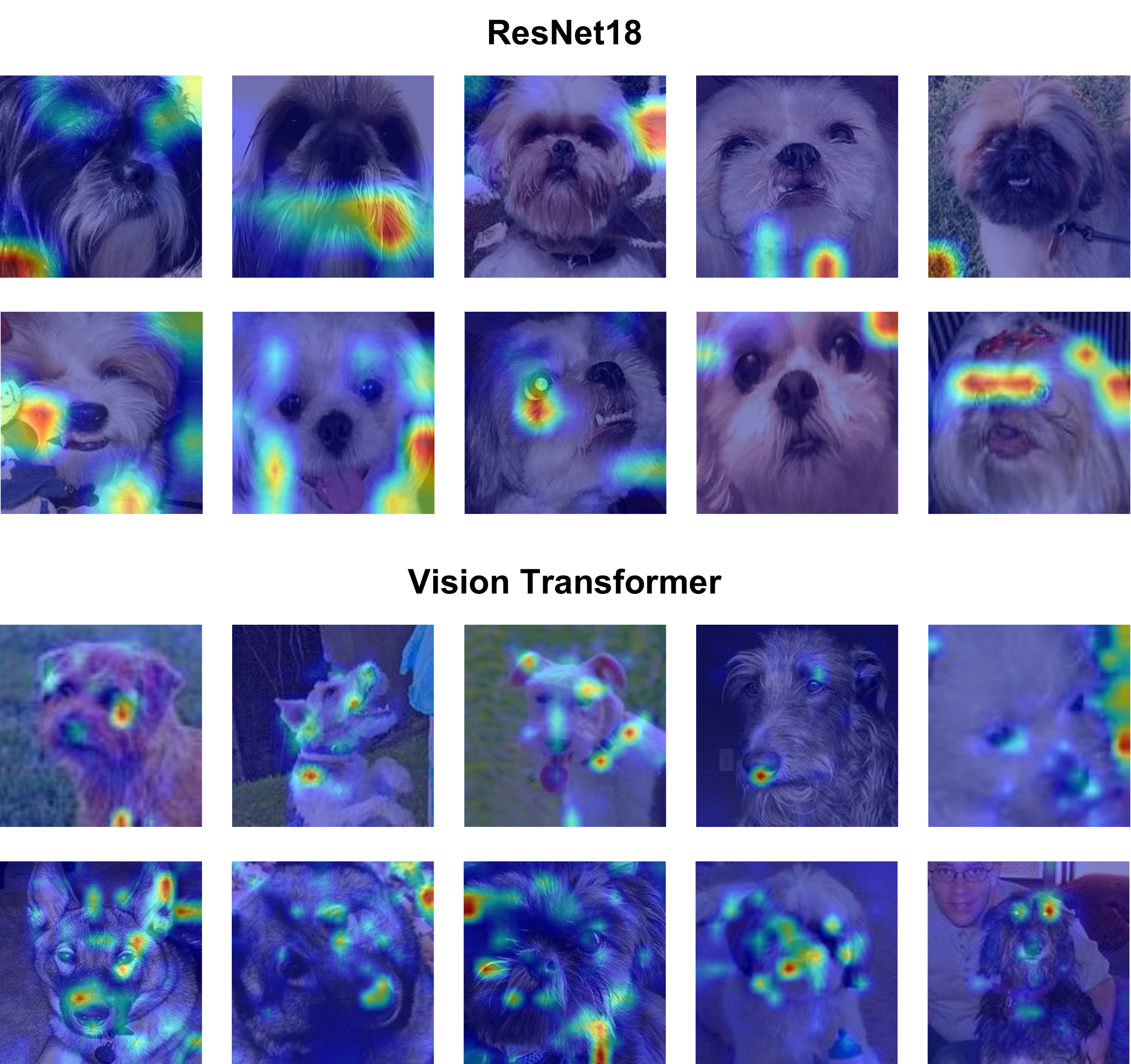}
	\caption{GradCam outputs of ResNet18 and Vision Transformer on random selected samples from Animal Faces.}
	\label{gcam}
\end{figure}

We performed GradCam visualization~\cite{gradcam} on the focus of the ResNet18 and ViT trained on the generated data.
Visualization in Fig.~\ref{gcam} shows that both classification models mainly focus on the information on some areas that may be ignored by humans.
In individual cases, ResNet18 even erroneously focuses only on the background information as the basis for discrimination.
This can partially explain why WaveGAN-generated data can maintain on-par classification results even though the overall collapse occurs, since its local generation is directly fused from the real data as a fusion-based method. 
The unsatisfying improvement of SAGE on downstream classifications may be attributed to the generalization ability of the neural classifier.

\begin{figure}[t]
	\centering
	\includegraphics[width=\linewidth]{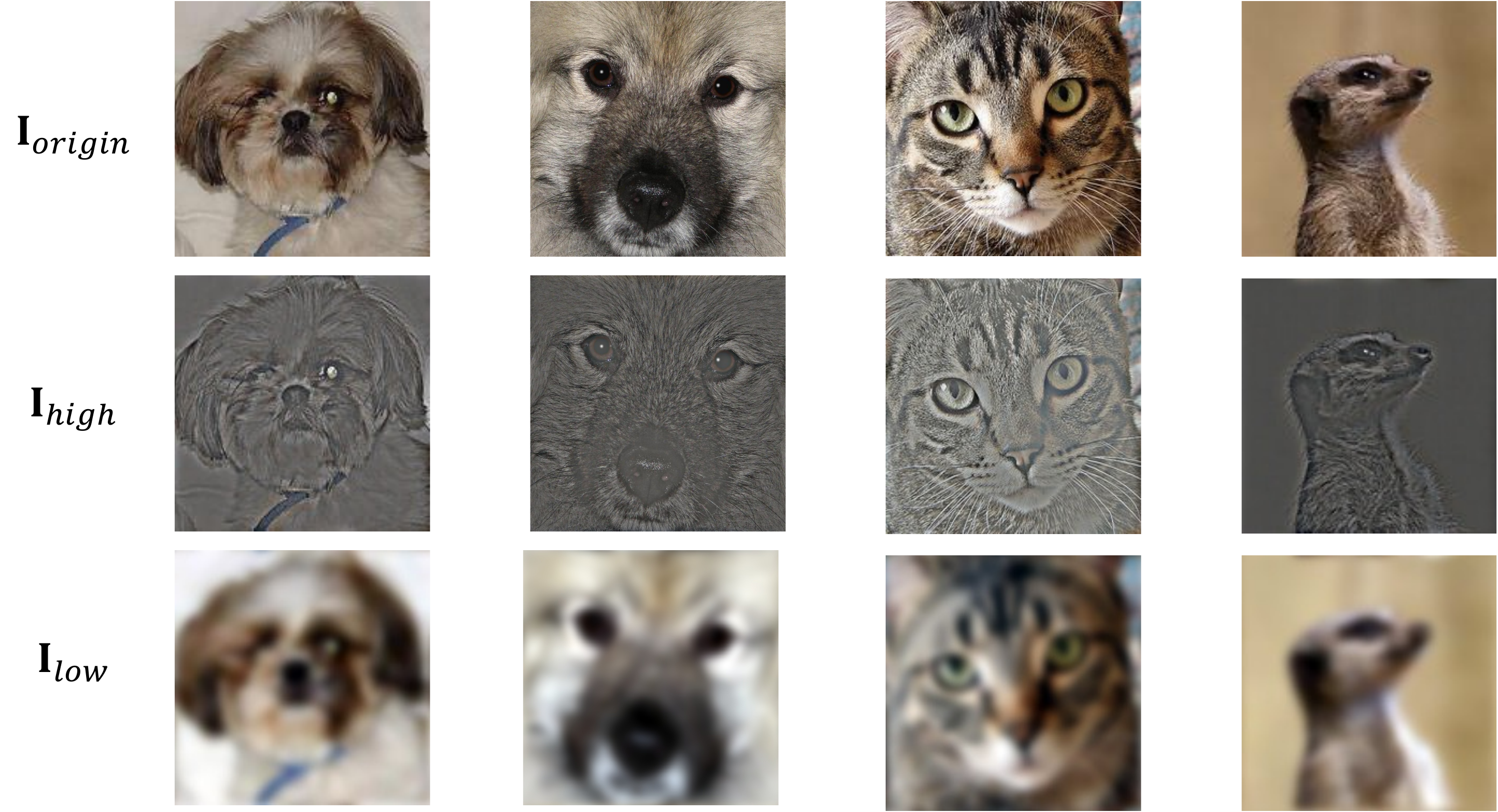}
	\caption{The images in the test split are decomposed into low-frequency parts and high-frequency parts.}
	\label{frequency_decompose}
\end{figure}

Additionally, we analyze the classifier sensitivity in the frequency domain. 
We decompose the test dataset of Animal Faces into low-frequency parts and high-frequency parts as shown in Fig.~\ref{frequency_decompose}.
Then a well-trained ViT is tested on the two parts of data respectively.
As the experimental results in TABLE~\ref{frequency_difference_test}, the classifier has much higher accuracy on low-frequency images.
This indicates that the classifier relies more on low-frequency information as the basis of discrimination.
In conclusion, the classifier pays more attention to some local structural features of the object during the classification process.
This information is lost in the inversion and generation process as demonstrated in Section~\ref{analyse_GAN}.

\begin{table}[t]
\renewcommand\arraystretch{1.5}
\centering
\caption{Classification accuracy of a well-trained classifier tested on low-frequency and high-frequency data on Animal Faces.}
\label{frequency_difference_test}
\begin{tabular}{lccc}
\hline
Frequency & High    & Low     & All               \\ \hline
Accuracy  & 17.93\%  & 44.58\%  & 77.96\% \\ \hline
\end{tabular}
\end{table}

\subsection{Solutions}
Based on the above experimental analysis, the most direct solutions to the class inconsistency problem are in the classification step and generation step.
The reason why the classifier only focuses on the local structure or even the background is that the data sparsity causes the model to rely too much on some biases, which is inconsistent with human visual perception mechanism.
Therefore, to solve this problem from the classifier perspective, we should work on enlarging the perceptual field and generalization ability of the classification model to alleviate over-fitting.
From the generation perspective, the image generation model should maintain the local structure of the objects so that the misalignment between inverse embedding and real embedding can be reduced. However, according to TABLE~\ref{inversion_comparison}, there is a trade-off in all the methods, {\it i.e.}, the stronger the ability to restore the image, the less editable the generation process is.
A better GAN inversion method that guarantees both editability and fidelity will definitely further improve the performance of SAGE. This conclusion can also extend to other generative models like VAEs~\cite{kingma2013auto,razavi2019generating} and Diffusion models~\cite{glide}, which inspires the development of generative augmentation model by maintaining the quality of low-frequency and local detail information.

However, it is hard to guarantee both editability and fidelity for GAN inversion. Existing methods like Hyperstyle~\cite{hyperstyle} fine-tune the generator for higher fidelity, but this strategy inevitably hurts the disentangling property of StyleGAN's latent space. 
On the other hand, robust neural classification is not the focus of this paper.
Upon SAGE and ViT, we provide two complementary tricks that can improve the performance of downstream classification after image generation.
According to Section~\ref{analyse_CLF}, these two tricks can well compensate for the lost information in the pixel domain and the frequency domain.

\begin{figure}[t]
	\centering
	\includegraphics[width=\linewidth]{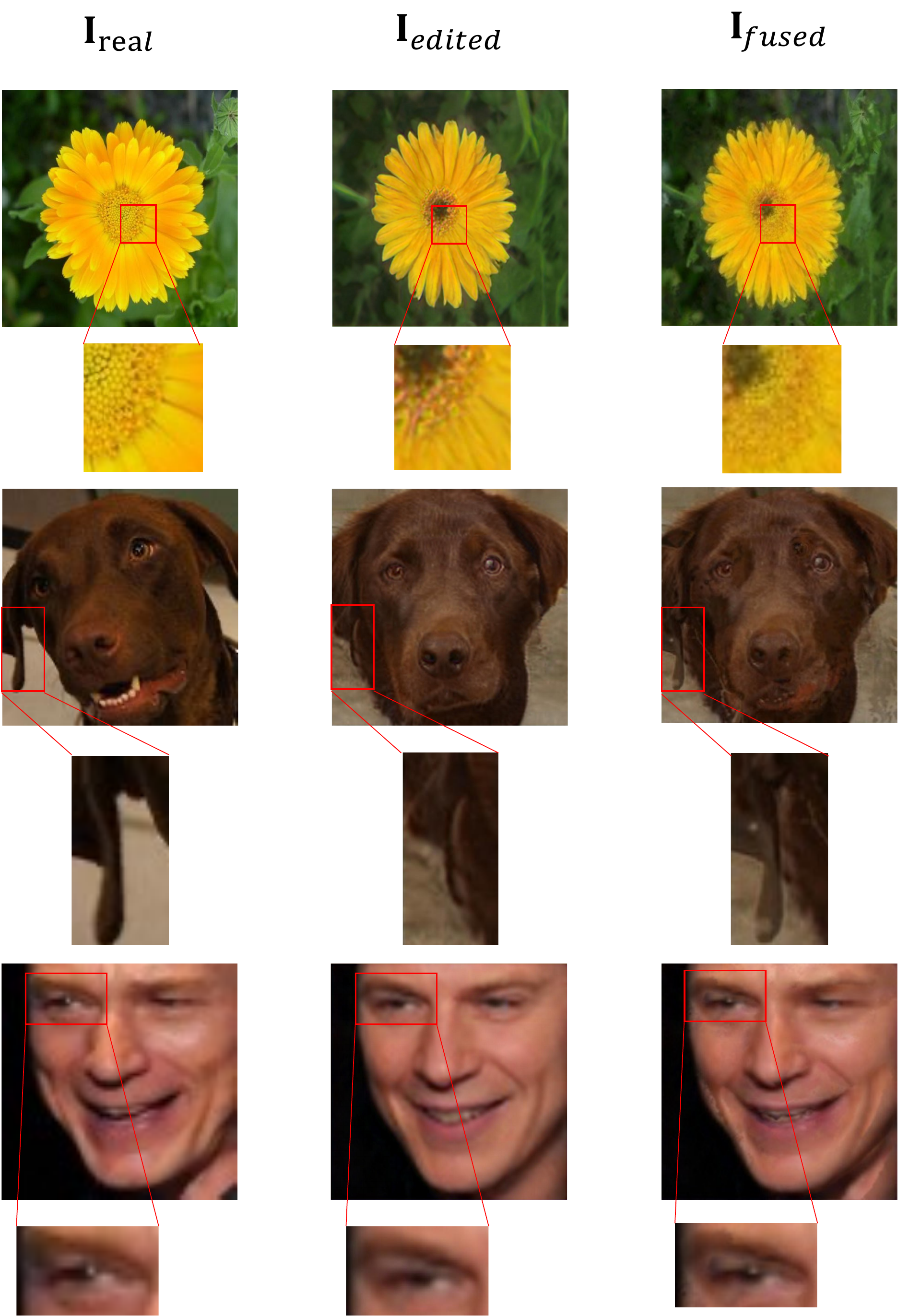}
	\caption{Visualization of the fused image generated by SAGE$_p$.}
	\label{mask_fusion}
\end{figure}

In the pixel domain, the generated images lack crucial local information that the classifiers focus on. Therefore, our first attempt is to directly add these features like fusion-based methods, named SAGE$_{p}$.
First, we generate a mask based on the difference between the edited image and the reconstructed image:
\begin{equation}
  \mathbf{Mask} = Gaussian(|| \mathbf{I}_{real}-\mathbf{I}_{inv}|-\beta | \mathbf{I}_{edited}-\mathbf{I}_{inv}||),
  \label{eq:}
\end{equation}
where $Gaussian(.)$ is a Gaussian filter that makes the fused image smoother and $\beta$ is the weighting factor.
In addition, the difference between the real image and the reconstructed image is also calculated, which is used to represent the loss in the inversion process.
This loss is added to the edited image according to the mask.
\begin{equation}                                        \mathbf{I}_{fused}=\mathbf{I}_{edited}+\mathbf{Mask} * \mathbf{I}_{real},
  \label{eq:}
\end{equation}
As shown in Fig.~\ref{mask_fusion}, some discriminative details are fused from the original images, \emph{e.g.}, the shape of ears in the second example. Despite not changing the whole image a lot, SAGE$_p$ can remarkably improve the classification accuracy on all datasets.

\begin{figure}[t]
	\centering
	\includegraphics[width=\linewidth]{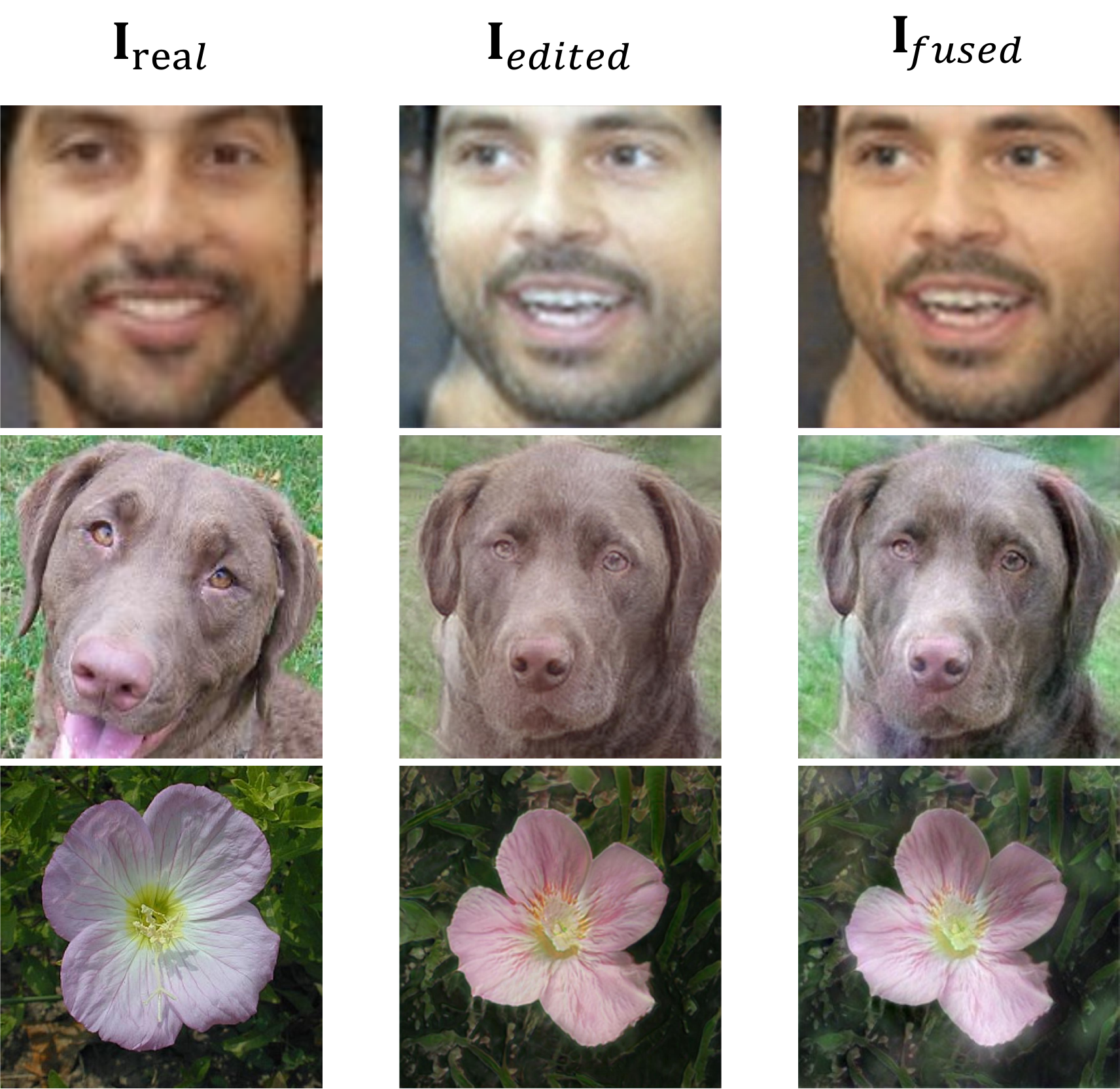}
	\caption{Visualization of the fused image generated by SAGE$_f$.}
	\label{frequency_fused}
\end{figure}

In the frequency domain, the major degradation comes from low frequency. Moreover, the neural classifier mainly focuses on the low-frequency information as analyzed in TABLE~\ref{frequency_difference_test}.
In SAGE$_f$, we add the difference between the low-frequency components of the real image and the reconstructed image to the generated image in the frequency space:
\begin{equation}   
\begin{split}
Low\_Pass&(\mathbf{I}_{fused})= \gamma_1 Low\_Pass(\mathbf{I}_{edited})\\
& +(Low\_Pass(\mathbf{I}_{real})-\gamma_2 Low\_Pass(\mathbf{I}_{inv})),
\end{split}
  \label{eq:}
\end{equation}
where $\gamma_1$ and $\gamma_2$ are the weight factors and $Low\_Pass$ indicates low-pass filtering. As shown in Fig.~\ref{frequency_fused}, low-frequency information changes the contour and the color of the generated objects. Compared with SAGE$_p$, the fusion in SAGE$_f$ is more natural to the naked eyes and achieves more obvious accuracy improvements on Animal Faces (58.98\%) and VGGFaces (77.13\%).

The function of SAGE$_p$ and SAGE$_f$ varies across different datasets. In Flowers, the hue or the counters makes few differences for different spices. SAGE$_f$ results in even lower accuracy. For VGGFaces, the hue of the images largely influences the people's skin color, and the contour of the face is crucial for recognition. Therefore, SAGE$_f$ makes much more improvements on this dataset.
Furthermore, we combine the implementation of both the pixel domain and the frequency domain. SAGE$_{p+f}$ achieves the best performance on all datasets, surpassing SAGE by 1.95\% on Animal Faces, 2.19\% on Flowers, 8.76\% on VGGFaces, and  3.20\% on NABirds. Moreover, the original SAGE underperforms ``Standard" on Flowers VGGFaces and NABirds where the discrepancy between different categories is small. With the two complementary techniques, SAGE$_{p+f}$ achieves an improvement over ``Standard'' on all the three datasets.

In conclusion, SAGE$_p$ and SAGE$_f$ straightforwardly complement the information with the classifier and SAGE unchanged. It can further support our analysis in the previous section and the rationality of the potential solutions. We hope these discussions can inspire further research on both robust neural classification and generative category retention.

\begin{table}[]
\renewcommand\arraystretch{1.5}
\centering
\caption{The Naive Augmentation Score of two complementary tricks.}
\label{attempts}
\begin{tabular}{lcccc}
\hline
Method   & Animal Faces     & Flowers          & VGGFaces*    & NABirds*    \\ \hline
Standard     & 55.38\%          & 79.41\%          & 72.13\%   & 52.60\%       \\ 
Inversed      & 49.57\%          & 52.64\%          & 52.22\%     & 36.42\%     \\ \hline
SAGE     & 57.37\%          & 77.82\%               & 70.53\% & 45.62\%   \\ 
SAGE$_p$ & 57.51\%          & 78.28\%          & 73.12\%     & 53.29\%     \\
SAGE$_f$  & 58.98\%          & 74.12\%          & 77.13\%     & 51.27\%    \\
SAGE$_{p+f}$  & 59.32\%          & 80.01\%          & 79.29\%     & 55.80\%    \\ \hline
\end{tabular}
\end{table}
\section{Conclusion}
In this paper, we investigate in-depth into the generative class inconsistency problem and propose Stable Attribute Group Editing (SAGE) for more robust class-relevant image generation. SAGE relocates the whole distribution of the novel class in the latent space, thus effectively avoids category-specific editing and enhances the generation stability. Experimental results show that SAGE can not only generate images with high quality but also better facilitate downstream tasks.
Going a step further, we find that class inconsistency is a common problem in generative models. We systematically discuss this issue from both the generative model and classification model perspectives. Extensive experiments are conducted to provide valuable insights on applying image generation to a wider range of downstream applications.
Along this direction, future research can be undertaken on enhancing the generative data augmentation from the perspectives of trustworthy downstream models (beyond classification) and the generation models (including GAN, Diffusion model and others).

\end{document}